\ifdefined\supponly\else\documentclass[10pt,twocolumn,letterpaper]{article}\fi

\usepackage[pagenumbers]{cvpr}      

\newcommand{\mace}{\textup{MACE}\xspace}

\newcommand{\macesb}{\ensuremath{\text{\mace}_{\text{SB}}}}
\newcommand{\sss}{\textit{S}\textsuperscript{3}}  

\definecolor{cvprblue}{rgb}{0.21,0.49,0.74}
\usepackage{xr-hyper}
\usepackage[pagebackref,breaklinks,colorlinks,allcolors=cvprblue]{hyperref}
\makeatletter
\newcommand*{\addFileDependency}[1]{\typeout{(#1)}\@addtofilelist{#1}\IfFileExists{#1}{}{\typeout{No file #1.}}}
\makeatother

\usepackage{amsmath,amssymb,bm}
\usepackage{algorithm}
\usepackage{algpseudocode}
\usepackage{graphicx}
\usepackage{booktabs}
\usepackage{multirow}
\usepackage{graphicx}
\usepackage{xspace}

\def\confName{CVPR}
\def\confYear{2027}

\title{MACE: Targeted Visual Counterfactual Explanations for\\
Contrastive Vision-Language Models}

\author{
Van Bach Nguyen$^{1}$ \qquad
Jörg Schlötterer$^{2}$ \qquad
Christin Seifert$^{1}$\\
$^{1}$University of Marburg, Germany\\
$^{2}$Hochschule RheinMain, Germany
}
\begin{document}
\ifdefined\supponly
\clearpage
\maketitlesupplementary

\appendix
\section{Additional Method Details}
\label{sec:appendix_method_details}

\subsection{Model-Specific Generative Processes}
\label{sec:method:backbones}

We detail how Stable Diffusion and I$^2$SB instantiate the common guidance and region-preservation procedure in the main paper. References to its shared update and restoration equations point back to the main paper. In both models, $D$ also maps generator outputs from $[-1,1]$ to $[0,1]$ and resizes them to the resolution of $x$. Since the sampler uses a subset of the diffusion timesteps, $t-1$ denotes the next timestep of the sampling schedule.

\paragraph{Latent diffusion (MACE-SD).}
We use a pretrained latent diffusion inpainting model~\citep{rombach2022latent} with VAE encoder $E$ and decoder $D$, where $E$ resizes its input to the generator's resolution and maps it from $[0,1]$ to $[-1,1]$ before encoding. Let $z_I=E(x)\in\mathbb R^{h\times w\times C_z}$ be the input latent, where $h\times w$ and $C_z$ are its spatial resolution and number of channels. Using the standard diffusion forward process~\citep{ho2020ddpm,rombach2022latent}, we construct the forward-noised reference trajectory
\begin{equation*}
z_I^{(t)}=\sqrt{\bar\alpha_t}\,z_I+
\sqrt{1-\bar\alpha_t}\,\varepsilon_I,\qquad
\varepsilon_I\sim\mathcal N(0,I),
\end{equation*}
where $t\in\{0,\ldots,T\}$ is the diffusion timestep, $\alpha_t=1-\beta_t$, $\bar\alpha_t=\prod_{s=1}^t\alpha_s$, and $\{\beta_t\}_{t=1}^T$ is the variance schedule, with $\bar\alpha_0=1$. The standard Gaussian noise $\varepsilon_I$ is sampled once and reused across timesteps. Following SDEdit~\citep{meng2022sdedit}, reverse diffusion starts from the noised input itself, $z_{t_1}=z_I^{(t_1)}$, so that the masked region keeps the coarse layout and colors of $x$.

At timestep $t$, the denoiser is conditioned on the target prompt, the latent mask $M_z$, obtained by nearest-neighbor downsampling of $M$ to $h\times w$, and the masked-image latent $z_{\mathrm{masked}}=E((1-M)\odot x+M\odot\tfrac12)$, in which the masked pixels are set to mid-gray, with classifier-free guidance~\citep{ho2022classifierfree} of scale $\omega$ against an empty prompt:
\begin{equation*}
\begin{aligned}
z_t^{\mathrm{in}}
&=\operatorname{concat}_{\mathrm{ch}}
  (z_t,M_z,z_{\mathrm{masked}}),\\
\widehat\varepsilon_t
&=\varepsilon_\theta(z_t^{\mathrm{in}},t,\bm h_{\varnothing})
 +\omega\big(\varepsilon_\theta(z_t^{\mathrm{in}},t,\bm h_{y_t})
 -\varepsilon_\theta(z_t^{\mathrm{in}},t,\bm h_{\varnothing})\big),
\end{aligned}
\end{equation*}
where $z_t^{\mathrm{in}}$ is the denoiser input, $\operatorname{concat}_{\mathrm{ch}}$ concatenates tensors along the channel dimension, and $\varepsilon_\theta$ is the pretrained denoiser. $\bm h_{y_t}$ is the diffusion model's conditioning embedding of the target prompt $t_{y_t}$ defined in Section~\ref{sec:problem}, and $\bm h_{\varnothing}$ is that of the empty prompt. The predicted clean latent is
\begin{equation*}
\hat z_0(z_t)=\frac{z_t-\sqrt{1-\bar\alpha_t}\,\widehat\varepsilon_t}
{\sqrt{\bar\alpha_t}}.
\end{equation*}
We set $q_t=\widehat\varepsilon_t$ and use the deterministic DDIM step~\citep{song2021ddim}, $\mathcal S_t(z_t,\widehat\varepsilon_t)=\sqrt{\bar\alpha_{t-1}}\,\hat z_0(z_t) +\sqrt{1-\bar\alpha_{t-1}}\,\widehat\varepsilon_t$. The guidance enters the shared update in \cref{eq:update} with the weight of the clean latent in this step:
\begin{equation}
\bar z_{t-1}=\mathcal S_t(z_t,\widehat\varepsilon_t)
+\lambda\underbrace{\sqrt{\bar\alpha_{t-1}}}_{w_t^{\mathrm{SD}}}\frac{\hat g_t}{\kappa},
\label{eq:sd-injection}
\end{equation}
that is, it acts as a correction $\lambda\hat g_t/\kappa$ of $\hat z_0(z_t)$. Protected latent locations are restored with \cref{eq:restore}, following the noised input trajectory~\citep{lugmayr2022repaint}. Since VAE decoding couples spatial locations, this preserves the protected latents but not the protected pixels exactly; the final composite in \cref{sec:counterfactual-inpainting} restores them.

\paragraph{I$^2$SB (MACE-SB).}
We use an image-to-image Schr\"odinger bridge~\citep{liu2023i2sb}, following \citep{sobieski2024rethinking}. Generation operates directly in image space on the square crop of $x$ that CLIP sees, resized to the bridge's resolution, so $z_t$ is an image, $D$ only rescales the values and places the crop back into $x$, and $M_z$ is $M$, cropped and resized in the same way as $x$, broadcast across image channels. The bridge transports the corrupted input $x_1=(1-M)\odot x+M\odot\varepsilon$, $\varepsilon\sim\mathcal N(0,I)$, to a clean image. Sampling starts from the deterministic bridge interpolation $z_{t_1}=\mu_0(t_1)\,x+\mu_1(t_1)\,x_1$ at the intermediate step $t_1$, with $\mu_0(t)=\bar V_t/(V_t+\bar V_t)$ and $\mu_1(t)=V_t/(V_t+\bar V_t)$, where $V_t$ and $\bar V_t$ are the bridge's cumulative noise variances from the clean end to $t$ and from $t$ to the corrupted end~\citep{liu2023i2sb}. The denoiser predicts the clean image $\hat z_0(z_t)$. For the deterministic mean update used by RCSB, we set $q_t=\hat z_0(z_t)$ and $\mathcal S_t(z_t,q_t)=a_t\hat z_0(z_t)+b_tz_t$, with $a_t=1-b_t$ and $b_t=V_{t-1}/V_t$, where $V_0=0$. As in RCSB, the guidance shifts the state before this step while the clean-image estimate, through which the guidance gradient was computed, is held fixed. Since the step is linear in the state, this is the shared update in \cref{eq:update} with
\begin{equation}
\bar z_{t-1}=a_t\hat z_0(z_t)+b_t\Big(z_t+\lambda\frac{\hat g_t}{\kappa}\Big)
=\mathcal S_t(z_t,\hat z_0(z_t))+
\lambda\underbrace{b_t}_{w_t^{\mathrm{SB}}}\frac{\hat g_t}{\kappa}.
\label{eq:sb-injection}
\end{equation}
For region preservation, we use $z_I^{(t)}=x$ at protected pixels in \cref{eq:restore}. Equations~\eqref{eq:sd-injection} and~\eqref{eq:sb-injection} show how the same normalized shift yields different effective guidance weights under the two generation processes.

\section{Experimental Setup Details}
\label{sec:appendix_setup}

\subsection{Dataset Sampling and Source-Target Pair Construction}
\label{sec:appendix_dataset_sampling}

For ImageNet, we score at most 20 validation images per class and select the first eligible image. Classes with no eligible image among those 20 are omitted, yielding 934 source-target pairs. For Food-101, we score 20 images per class and select at most 10 eligible images per class, yielding 1,003 pairs. For Oxford-IIIT Pet, we score the full test split and select at most 35 images per class, yielding 1,194 pairs.

For CUB-200-2011, we score all 5,794 test images against the 200 bird classes and select at most 10 correctly classified images per class. This yields 1,595 pairs from 186 source classes. Ablations use a separate 100-instance subset, with one example from each of 100 source classes.

\subsection{Implementation Details}
\label{sec:appendix_implementation}

The hyperparameters of both \mace variants are shared across all datasets.

\paragraph{Adaptive mask expansion.} Both variants use the same mask schedule. The attribution maps are computed with CAV~\cite{chen2025class} and combined into $R_{\mathrm{diff}}$ (\cref{eq:contrastive-relevance}). The mask fraction grows as $r_k=0.05k$ for $k=1,\dots,K$ with $K=20$, i.e., $r_k=0.05, 0.10, \dots, 1.00$, and each mask is dilated by 10 pixels. The expansion stops at the first candidate whose top-1 prediction is the target; if no candidate succeeds, it returns the candidate with the largest target margin.

\paragraph{Generator.} \mace{} uses Stable Diffusion~1.5 inpainting at $512\times512$ with a DDIM sampler (50 steps), starting the reverse process at $t_1=0.7\,T$ (SDEdit~\cite{meng2022sdedit}). It is conditioned on the bare target class name with classifier-free guidance scale 7.5, and uses seed 0. \macesb{} uses I$^2$SB with the RCSB checkpoint on a $256\times256$ centre crop, with the bridge mean update and 100 function evaluations, starting at $t_1=0.7\,T$ without text conditioning, and uses seed 42.

\paragraph{Guidance.} Both variants minimize the CLIP cross-entropy over the class vocabulary (\cref{eq:clip-ce}) and stabilize the gradient with Adam moments, normalized by the norm of the first stabilized gradient (\cref{eq:stabilization}). For every mask size, \mace{} tries the guidance strengths $\lambda=6$. \macesb{} generates once per mask size with RCSB's guidance strength of 1.5.

\subsection{Adaptation of the Compared Methods}
\label{sec:appendix_baselines}

All compared methods use their official implementation. The change shared by all of them is the explained model: the supervised classifier is replaced by the zero-shot classifier of the explained CLIP, whose logits are the cosine similarities between the image embedding and the text embeddings of all class names of the dataset, multiplied by 100. The class prompts are the same as for \mace. Each method runs at its own resolution on a centre crop of the image, and the result is mapped back onto the original image at its native resolution for evaluation.

\paragraph{LDCE.} We use the text-conditional variant with the hyperparameters of the original paper: the ImageNet setting for ImageNet and the Oxford-IIIT Pet setting, which the authors use for a CLIP classifier, for the other three datasets. The diffusion model is conditioned on the target class with the prompt templates of the original paper, and with analogous templates for Food-101 and CUB-200-2011.

\paragraph{DiG-IN.} We use the setting of the original paper for counterfactual generation with Stable Diffusion~1.4. The image is resized and centre-cropped to $512\times512$ (the official implementation resizes to 640 before cropping, which leaves the border of the image unedited), so that the method sees the same region as the other methods. Instead of a generated caption, the inversion uses the caption ``an image of a \emph{source class} on background''.

\paragraph{MaskDiME.} The released code covers only CelebA. We therefore reimplement the ImageNet pipeline from the released components with the hyperparameters of the original paper, using the unconditional $256\times256$ ImageNet diffusion model of Dhariwal and Nichol~\cite{dhariwal2021diffusion} for all datasets.

\paragraph{RCSB.} We use the automated region selection with the setting that the original paper reports for CLIP (area $0.2$, cell size $16$, guidance scale $1.5$, trajectory truncation $0.7$) and the released I$^2$SB checkpoint for all datasets.

\section{Additional Experimental Results}
\raggedbottom
\label{sec:appendix}

This section reports the results of the pixel-space variant \macesb{}, the comparison with lower-ranked and very low-ranked target classes, the effect of the generator's text prompt, the counterfactual transition, the comparison with other explained models and other attribution methods, additional qualitative comparisons, the distance of the counterfactuals past the decision boundary, examples of counterfactuals with increasing mask size, examples of the same-mask and same-size experiments, and the normalization of the guidance step.

\subsection{Pixel-Space Variant \macesb{}}
\label{sec:appendix_macesb}

\Cref{tab:appendix_main_full} repeats the comparison of \cref{tab:main_results} with the pixel-space variant \macesb{} included, and \cref{tab:appendix_variants} compares \mace and \macesb{} directly. Both reach at least 99.9\% validity on every dataset. \macesb{} makes slightly smaller edits, with the lowest $\ell_1$, $\ell_2$, and LPIPS and the highest SSIM on all four datasets, while \mace{} achieves the lower FID and sFID on all four datasets. The differences between \mace and \macesb{} are small compared to their distance to the other methods, so the results of \mace do not depend on the generative model. \Cref{fig:appendix_qual_macesb} adds \macesb{} to the qualitative comparison in \cref{fig:qualitative_all_datasets}: on all three examples, \macesb{} edits a similar region as \mace{} and also reaches the target.

\subsection{Lower-Ranked Target Classes}
\label{sec:appendix_target_rank}

The main evaluation uses the class ranked second by CLIP as the target. Tables~\ref{tab:appendix_rank5} and~\ref{tab:appendix_rank10} repeat the comparison with harder targets, the classes ranked 6th and 11th, for the same source images and with the same metrics and protocol as Table~\ref{tab:main_results}.

To test much harder targets, we take the class at about 50\% and 70\% of CLIP's ranking (e.g., rank 500 and 700 of 1000 for ImageNet) as the target, on the full benchmark of each dataset and with the same settings as for the 6th- and 11th-ranked targets; DiG-IN is not included in this comparison. \Cref{tab:appendix_rank_low50,tab:appendix_rank_low70} show that \mace{} and \macesb{} remain valid on at least 99.5\% of the instances, while the validity of the baselines drops further (MaskDiME to 0.45 and LDCE to 0.66 on ImageNet). \macesb{} makes the smallest changes by $\ell_1$, LPIPS, and SSIM in every setting and \mace{} has the second-lowest LPIPS, although the edits grow with the distance between source and target class (LPIPS of \mace{} 0.045--0.067, compared with 0.029--0.038 for the 6th-ranked target). In realism, \mace{}, \macesb{}, and MaskDiME, which all edit only inside a mask, reach the three lowest FID and sFID values on Food-101, Oxford Pets, and CUB-200; MaskDiME has the lowest FID in most settings but is valid on only 45--82\% of the instances. On ImageNet, where several baselines fail often, the common valid set is small (224--230 instances, compared with 346--437 for the 6th- and 11th-ranked targets), which inflates FID and sFID.

\subsection{Text Prompt of the Generator}
\label{sec:appendix_no_prompt}

\mace{} conditions Stable Diffusion on the bare target class name. To test how much this text condition contributes, we replace it with an empty prompt and keep CLIP guidance, mask construction, and all other settings unchanged. \Cref{tab:appendix_no_prompt} shows that the text prompt has almost no effect: on the full benchmarks, both variants are valid on at least 99.9\% of the instances and reach nearly the same proximity (LPIPS 0.027 vs.\ 0.028). Without the prompt, the smallest mask suffices slightly more often (91.2\% vs.\ 88.1\%), the search needs slightly fewer generations (1.81 vs.\ 1.98), and CMV is slightly lower (0.094 vs.\ 0.104). The edits are therefore driven by the CLIP guidance rather than by the text condition of the generator.

\subsection{Other Explained Models}
\label{sec:appendix_target_models}

We repeat the comparison with OpenAI CLIP ViT-B/32 and ViT-L/14 and SigLIP ViT-B/16 as the explained model. Source images, targets, and evaluation follow the main protocol, with the respective model used for source selection, guidance, and validity, on the full benchmark of each dataset (926--1868 instances per model and dataset). Tables~\ref{tab:appendix_target_l14}--\ref{tab:appendix_target_siglip} report the results. For all three models, \macesb{} is valid on at least 99.8\% of the instances for every dataset, and \mace{} on at least 99.9\% for the OpenAI models (0.93--0.997 for SigLIP), the highest or near-highest validity. \mace{} or \macesb{} makes the smallest changes (lowest LPIPS and highest \sss{} in every setting) and reaches the lowest FID in 8 of 12 settings, with MaskDiME close behind. Their cross-model validity remains low (0.08--0.19 on average over the datasets), as for ViT-B/16 (\cref{sec:cmv}), whereas it reaches 0.74--0.81 for Stable Diffusion and DiG-IN.

\subsection{Other Attribution Methods}
\label{sec:appendix_attribution}

We replace CAV with Grad-ECLIP~\cite{zhao2024gradient}, LeGrad~\cite{bousselham2025legrad}, and M2IB~\cite{wang2024m2ib} to compute the attribution maps $A_s$ and $A_t$ in \cref{eq:contrastive-relevance}, keeping the rest of \mace unchanged. Since CAV is designed for OpenAI CLIP, we compare it with Grad-ECLIP on ViT-B/16, ViT-L/14, and ViT-B/32, additionally with M2IB on ViT-B/16, and compare LeGrad with Grad-ECLIP on SigLIP. All comparisons use the full benchmark of each dataset. \Cref{tab:appendix_attribution} shows that the attribution method barely affects \mace on the OpenAI CLIP models, neither in proximity nor in realism (FID, KID, sFID). M2IB is slightly worse than CAV: its source and target maps are nearly identical for related classes, so the difference map leaves only a small region and the mask search needs more attempts, which lowers validity and CMV slightly. On SigLIP, LeGrad yields clearly higher validity than Grad-ECLIP for \mace{} (0.97 vs.\ 0.80), so the attribution must highlight regions that are relevant to the explained model's decision.

\subsection{Fixed Mask Sizes}
\label{sec:appendix_fixed_mask}

\Cref{tab:appendix_fixed_mask} extends the mask-size ablation of \cref{tab:mask_ablation} to all evaluated mask sizes. Validity rises quickly with $r$ and reaches 1.0 at $r=0.5$, while all proximity metrics degrade steadily with the mask area. The adaptive search reaches full validity at the proximity of the smallest mask ($r=0.05$), since it enlarges the mask only for the instances that need it.

\subsection{Additional Qualitative Comparisons}
\label{sec:appendix_qualitative}

\Cref{fig:appendix_qual_a} extends the qualitative comparison in \cref{fig:qualitative_all_datasets} to all four datasets. \Cref{fig:appendix_qual_a} shows instances with lower-ranked targets (ranked 6th or 11th by CLIP, \cref{sec:appendix_target_rank}) on which \mace{} performs well: its counterfactual changes a small region on the object (at most 6\% of the image), visibly adds target-class features, and is valid for at least four of the five held-out CLIPs. Stable Diffusion typically replaces the whole scene, LDCE alters the global appearance, DiG-IN also changes the whole image but concentrates the visible change on the object, MaskDiME spreads small changes over the image, and RCSB changes larger regions than \mace, whose edits remain confined to a few regions selected by the counterfactual mask.

\subsection{Distance Past the Decision Boundary}
\label{sec:appendix_boundary}

\mace's adaptive mask expansion stops at the first candidate for which the target becomes the top-1 prediction of the explained model. \Cref{fig:appendix_boundary}a shows the resulting target probability of the valid counterfactuals under the explained CLIP. \mace{}'s counterfactuals lie just past the decision boundary, with a median target probability of 0.64, whereas LDCE and RCSB reach a median of 0.99. \Cref{fig:appendix_boundary}b relates this distance to cross-model validity. For every method, CMV increases with the target probability under the explained model. For \mace{}, CMV rises from 0.07 for counterfactuals with a target probability below 0.5 to 0.36 for those between 0.95 and 0.99. Counterfactuals that only just cross the explained model's decision boundary therefore rarely cross the boundaries of other models (\cref{sec:cmv}). The distance past the boundary is not the only factor, however: \macesb{} stops further past the boundary (median 0.95) but reaches a CMV similar to \mace{}, and at the same target probability, methods that change larger regions, such as Stable Diffusion and DiG-IN, transfer much more.

\subsection{Same Mask Size}
\label{sec:appendix_same_size}
Besides restricting \mace{} to the regions changed by each compared method (same mask, \cref{sec:cmv}), we also run \mace{} with masks with the same area as each compared method's edited region, selected using \mace{}'s counterfactual mask (same size). Adaptive expansion is disabled in both settings. \Cref{fig:appendix_same_size} and \cref{tab:appendix_same_size} show the results. With the same mask size, \mace{} exceeds the CMV of LDCE (0.59 vs.\ 0.44) and MaskDiME (0.45 vs.\ 0.12) and matches Stable Diffusion (0.74) at a much lower LPIPS, but remains below DiG-IN (0.58 vs.\ 0.82) and RCSB (0.21 vs.\ 0.25). For the region of every compared method, its CMV is lower than with the same mask.

\subsection{Counterfactuals with Increasing Mask Size}
\label{sec:appendix_size_examples}

\Cref{fig:appendix_size_examples} illustrates how \mace{}'s counterfactuals change as the editable region grows, for one selected instance per dataset. Here, the mask is fixed to the top $r$ fraction of $R_{\mathrm{diff}}$ instead of being chosen by the adaptive search, and each mask size is generated separately. Small masks change only a few regions and are valid mostly for the explained model (CMV 0.0--0.2). As the mask grows, \mace{} adds more target-class evidence, such as the donut shape and sprinkles for Food-101, the black-and-tan face of the Miniature Pinscher for Oxford Pets, or the reddish head of the Horned Grebe for CUB-200, and more held-out models predict the target.

\subsection{Same Mask and Same Mask Size: Examples}
\label{sec:appendix_region_examples}

\Cref{fig:appendix_region_imagenet,fig:appendix_region_food101,fig:appendix_region_oxford_pets,fig:appendix_region_cub200} illustrate the same-mask experiment of \cref{sec:cmv} and the same-size experiment of \cref{sec:appendix_same_size} on one randomly selected instance per dataset. For the large regions of Stable Diffusion, LDCE, and DiG-IN, \mace{} keeps the scene layout and adds target evidence inside the region, and in most cases reaches a CMV similar to or higher than the compared method's, with a lower LPIPS than Stable Diffusion. For the small regions, the results vary: in MaskDiME's region, \mace{} reaches a higher CMV than MaskDiME in three of the four examples, whereas in RCSB's region it exceeds RCSB only with the same size on Food-101. On CUB-200, most variants remain specific to the explained model. These examples reflect the average trend in \cref{fig:cmv_vs_edit,fig:appendix_same_size}: given a region of comparable size, \mace{}'s counterfactuals transfer about as well as the compared methods'.

\subsection{Normalization of the Guidance Step}
\label{sec:appendix_l2}

\mace{} stabilizes the CLIP gradient with Adam moments (\cref{eq:stabilization}). The Adam update divides each latent element by its own gradient magnitude, so all elements move by a similar amount. As an alternative, we replace it by the $\ell_2$-normalized gradient $g_t/\|g_t\|_2$ at every denoising step, without momentum, which keeps the relative magnitudes of the gradient; the guidance strength and all other settings are unchanged. \Cref{tab:appendix_l2} compares both on the full benchmarks. Both are valid on at least 99.9\% of the instances. The $\ell_2$ step raises CMV from 0.104 to 0.155 on average, at slightly larger edits (LPIPS 0.032 vs.\ 0.026) and slightly lower realism (FID 36.9 vs.\ 35.3, sFID 75.0 vs.\ 73.7). It moves the counterfactuals further past the decision boundary of the explained model, which makes them transfer to other models more often. We keep the Adam step as the default, since it yields the smallest edits.

\Cref{tab:appendix_l2_region} repeats the comparison in the same-mask and same-size settings of \cref{sec:appendix_same_size}. With the $\ell_2$ step, CMV is higher by 0.04--0.11 in every region and at every size, at nearly the same LPIPS.

\clearpage
\begin{table*}[t]
    \centering
    \tiny
    \setlength{\tabcolsep}{3pt}
\resizebox{\textwidth}{!}{%
\begin{tabular}{lllccccccccc}
\toprule
 & & & \multirow{2}{*}{Validity $\uparrow$} & \multicolumn{5}{c}{Proximity} & \multicolumn{3}{c}{Realism} \\
\cmidrule(lr){5-9}\cmidrule(lr){10-12}
Dataset & Gen. & Method/Variant &
&
$\ell_1^{\text{norm}}\downarrow$ &
$\ell_2^{\text{norm}}\downarrow$ &
LPIPS $\downarrow$ &
SSIM $\uparrow$ &
\sss{} $\uparrow$ &
FID $\downarrow$ &
KID $\downarrow$ &
sFID $\downarrow$
\\
\midrule
\multirow{7}{*}{ImageNet}
 & Latent & Stable Diffusion & 0.7195 & 0.3337 & 0.4050 & 0.8378 & 0.1327 & 0.2165 & 102.0257 & 4.87 & 160.37 \\
 & Latent & LDCE & 0.9711 & 0.0514 & 0.0919 & 0.3263 & 0.6064 & 0.8266 & 84.8138 & 6.30 & 166.13 \\
 & Latent & DiG-IN & 0.9604 & 0.0547 & 0.0974 & 0.2765 & 0.7051 & 0.6571 & 78.2886 & 4.91 & \textbf{154.22} \\
 & Latent & \textbf{\mace{} (ours)} & \textbf{0.9989} & \underline{0.0045} & 0.0306 & \underline{0.0240} & \underline{0.9707} & \textbf{0.9938} & \textbf{75.6125} & 0.29 & \underline{159.28} \\
 & Pixel & MaskDiME & 0.7859 & 0.0133 & \underline{0.0287} & 0.1009 & 0.9068 & 0.9797 & \underline{76.0688} & \textbf{0.01} & 161.50 \\
 & Pixel & RCSB & 0.9957 & 0.0118 & 0.0460 & 0.0751 & 0.9191 & 0.9698 & 78.6671 & 1.56 & 162.06 \\
 & Pixel & \textbf{\macesb{} (ours)} & \textbf{0.9989} & \textbf{0.0033} & \textbf{0.0229} & \textbf{0.0237} & \textbf{0.9760} & \underline{0.9937} & 76.3184 & \underline{0.27} & 160.77 \\
\midrule
\multirow{7}{*}{Food-101}
 & Latent & Stable Diffusion & 0.9701 & 0.3418 & 0.4199 & 0.8188 & 0.1751 & 0.2875 & 149.5792 & 31.54 & 155.60 \\
 & Latent & LDCE & 0.9930 & 0.0436 & 0.0724 & 0.3134 & 0.6420 & 0.8238 & 33.9574 & 8.60 & 66.82 \\
 & Latent & DiG-IN & 0.9930 & 0.0517 & 0.0834 & 0.2449 & 0.7427 & 0.7300 & 28.3745 & 1.24 & 62.78 \\
 & Latent & \textbf{\mace{} (ours)} & 0.9990 & \underline{0.0044} & 0.0274 & \underline{0.0308} & \underline{0.9685} & \underline{0.9919} & \textbf{24.4550} & 1.30 & \textbf{55.85} \\
 & Pixel & MaskDiME & 0.8993 & 0.0094 & \textbf{0.0178} & 0.0871 & 0.9349 & 0.9840 & \underline{24.5360} & \underline{1.15} & 56.41 \\
 & Pixel & RCSB & \textbf{1.0000} & 0.0120 & 0.0425 & 0.0798 & 0.9143 & 0.9746 & 25.4409 & 1.77 & 57.70 \\
 & Pixel & \textbf{\macesb{} (ours)} & \textbf{1.0000} & \textbf{0.0035} & \underline{0.0218} & \textbf{0.0290} & \textbf{0.9713} & \textbf{0.9927} & 24.5821 & \textbf{1.07} & \underline{56.34} \\
\midrule
\multirow{7}{*}{Oxford Pets}
 & Latent & Stable Diffusion & 0.8970 & 0.3186 & 0.3906 & 0.8369 & 0.1656 & 0.2856 & 131.6522 & 20.14 & 135.99 \\
 & Latent & LDCE & 0.9791 & 0.0377 & 0.0671 & 0.2697 & 0.6606 & 0.8530 & 38.9011 & 6.77 & 66.27 \\
 & Latent & DiG-IN & 0.9841 & 0.0463 & 0.0836 & 0.2240 & 0.7492 & 0.6941 & 38.2877 & 5.70 & 66.70 \\
 & Latent & \textbf{\mace{} (ours)} & \textbf{1.0000} & \underline{0.0044} & 0.0300 & \underline{0.0248} & \underline{0.9707} & \underline{0.9901} & \underline{25.3779} & \textbf{1.11} & \underline{50.60} \\
 & Pixel & MaskDiME & 0.8157 & 0.0115 & \underline{0.0243} & 0.0775 & 0.9203 & 0.9825 & \textbf{24.7980} & \underline{1.52} & \textbf{49.91} \\
 & Pixel & RCSB & 0.9975 & 0.0112 & 0.0427 & 0.0610 & 0.9218 & 0.9631 & 32.1134 & 3.31 & 58.11 \\
 & Pixel & \textbf{\macesb{} (ours)} & \textbf{1.0000} & \textbf{0.0030} & \textbf{0.0212} & \textbf{0.0218} & \textbf{0.9763} & \textbf{0.9910} & 25.6341 & 1.73 & 51.11 \\
\midrule
\multirow{7}{*}{CUB-200}
 & Latent & Stable Diffusion & 0.5034 & 0.2856 & 0.3481 & 0.8363 & 0.1763 & 0.3012 & 63.8571 & 11.68 & 66.74 \\
 & Latent & LDCE & 0.9900 & 0.0328 & 0.0645 & 0.2186 & 0.7475 & 0.9185 & 21.8873 & 2.14 & 39.94 \\
 & Latent & DiG-IN & 0.9843 & 0.0310 & 0.0630 & 0.1463 & 0.8453 & 0.9240 & 22.4184 & 3.20 & 38.43 \\
 & Latent & \textbf{\mace{} (ours)} & \textbf{1.0000} & \underline{0.0045} & 0.0313 & \underline{0.0230} & \underline{0.9732} & \textbf{0.9959} & \textbf{15.5555} & \textbf{1.00} & \textbf{31.79} \\
 & Pixel & MaskDiME & 0.9254 & 0.0090 & \textbf{0.0216} & 0.0668 & 0.9435 & 0.9890 & 16.7839 & \textbf{1.00} & 33.69 \\
 & Pixel & RCSB & \textbf{1.0000} & 0.0122 & 0.0485 & 0.0719 & 0.9223 & 0.9680 & 25.4165 & 2.42 & 43.87 \\
 & Pixel & \textbf{\macesb{} (ours)} & \textbf{1.0000} & \textbf{0.0032} & \underline{0.0233} & \textbf{0.0222} & \textbf{0.9775} & \underline{0.9950} & \underline{16.7811} & 1.21 & \underline{33.16} \\
\bottomrule
\end{tabular}%
}
\caption{Quantitative comparison of counterfactual generation methods, including the pixel-space variant \macesb{}. We report target-class validity, proximity and realism metrics. All metrics except validity are computed on the common valid set. KID is reported in units of $10^{-3}$. Bold and underlined values indicate the best and second-best results, respectively, within each dataset and metric column. Gen. indicates the diffusion model type (latent or pixel space).}
\label{tab:appendix_main_full}
\end{table*}

\begin{table*}[t]
    \centering
    \small
    \setlength{\tabcolsep}{4pt}
\begin{tabular}{llccccccccc}
\toprule
 & & \multirow{2}{*}{Validity $\uparrow$} & \multicolumn{5}{c}{Proximity} & \multicolumn{3}{c}{Realism} \\
\cmidrule(lr){4-8}\cmidrule(lr){9-11}
Dataset & Variant & & $\ell_1^{\text{norm}}\downarrow$ & $\ell_2^{\text{norm}}\downarrow$ & LPIPS $\downarrow$ & SSIM $\uparrow$ & \sss{} $\uparrow$ & FID $\downarrow$ & KID $\downarrow$ & sFID $\downarrow$ \\
\midrule
\multirow{2}{*}{ImageNet} & \mace{} & \textbf{0.9989} & 0.0045 & 0.0306 & 0.0240 & 0.9707 & \textbf{0.9938} & \textbf{75.6125} & 0.29 & \textbf{159.28} \\
 & \macesb{} & \textbf{0.9989} & \textbf{0.0033} & \textbf{0.0229} & \textbf{0.0237} & \textbf{0.9760} & 0.9937 & 76.3184 & \textbf{0.27} & 160.77 \\
\midrule
\multirow{2}{*}{Food-101} & \mace{} & 0.9990 & 0.0044 & 0.0274 & 0.0308 & 0.9685 & 0.9919 & \textbf{24.4550} & 1.30 & \textbf{55.85} \\
 & \macesb{} & \textbf{1.0000} & \textbf{0.0035} & \textbf{0.0218} & \textbf{0.0290} & \textbf{0.9713} & \textbf{0.9927} & 24.5821 & \textbf{1.07} & 56.34 \\
\midrule
\multirow{2}{*}{Oxford Pets} & \mace{} & \textbf{1.0000} & 0.0044 & 0.0300 & 0.0248 & 0.9707 & 0.9901 & \textbf{25.3779} & \textbf{1.11} & \textbf{50.60} \\
 & \macesb{} & \textbf{1.0000} & \textbf{0.0030} & \textbf{0.0212} & \textbf{0.0218} & \textbf{0.9763} & \textbf{0.9910} & 25.6341 & 1.73 & 51.11 \\
\midrule
\multirow{2}{*}{CUB-200} & \mace{} & \textbf{1.0000} & 0.0045 & 0.0313 & 0.0230 & 0.9732 & \textbf{0.9959} & \textbf{15.5555} & \textbf{1.00} & \textbf{31.79} \\
 & \macesb{} & \textbf{1.0000} & \textbf{0.0032} & \textbf{0.0233} & \textbf{0.0222} & \textbf{0.9775} & 0.9950 & 16.7811 & 1.21 & 33.16 \\
\bottomrule
\end{tabular}
\caption{Comparison of \mace (latent diffusion) and its pixel-space variant \macesb{} (Schr\"odinger bridge). Validity is computed over all instances and the other metrics on the common valid set of Table~\ref{tab:appendix_main_full}. KID is reported in units of $10^{-3}$. Bold marks the better variant.}
\label{tab:appendix_variants}
\end{table*}

\begin{figure*}[t]
    \centering
    \includegraphics[width=\textwidth]{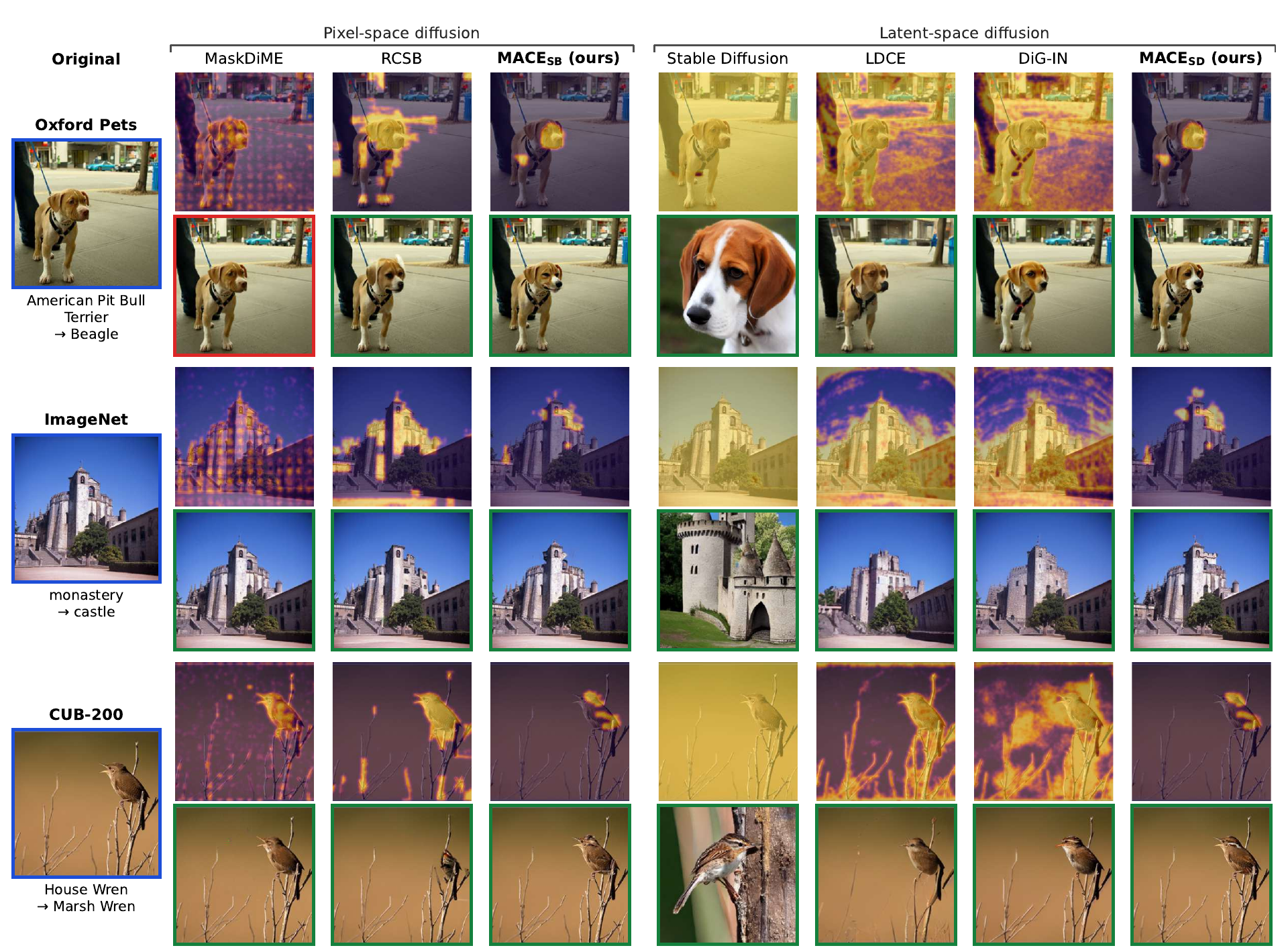}
    \caption{Qualitative comparison of Figure~\ref{fig:qualitative_all_datasets} with the pixel-space variant \macesb{} included. Layout as in Figure~\ref{fig:qualitative_all_datasets}.}
    \label{fig:appendix_qual_macesb}
\end{figure*}

\begin{table*}[t]
    \centering
    \small
    \setlength{\tabcolsep}{3pt}
\resizebox{\textwidth}{!}{%
\begin{tabular}{lllcccccccccc}
\toprule
 & & & Validity & \multicolumn{5}{c}{Proximity} & \multicolumn{3}{c}{Realism} \\
\cmidrule(lr){4-4}\cmidrule(lr){5-9}\cmidrule(lr){10-12}
Dataset & Gen. & Method/Variant & Target $\uparrow$ &
$\ell_1^{\text{norm}}\downarrow$ & $\ell_2^{\text{norm}}\downarrow$ & LPIPS $\downarrow$ & SSIM $\uparrow$ & \sss{} $\uparrow$ &
FID $\downarrow$ & KID $\downarrow$ & sFID $\downarrow$ \\
\midrule
\multirow{7}{*}{ImageNet}
& \multirow{4}{*}{Latent} & Stable Diffusion & 0.6991 & 0.3320 & 0.4033 & 0.8398 & 0.1295 & 0.0991 & 113.3189 & 5.75 & \textbf{171.25} \\
&  & LDCE & 0.9293 & 0.0530 & 0.0936 & 0.3351 & 0.5922 & 0.7781 & 100.7532 & 9.42 & 185.37 \\
&  & DiG-IN & 0.9143 & 0.0596 & 0.1047 & 0.3117 & 0.6816 & 0.5002 & 95.3340 & 12.08 & \underline{173.62} \\
&  & \mace{} (ours) & \textbf{1.0000} & \underline{0.0058} & 0.0360 & \underline{0.0325} & \underline{0.9637} & \underline{0.9878} & 88.1086 & \textbf{0.58} & 174.36 \\
& \multirow{3}{*}{Pixel} & MaskDiME & 0.7088 & 0.0139 & \underline{0.0295} & 0.1060 & 0.9033 & 0.9797 & \underline{87.8248} & 1.19 & 175.38 \\
&  & RCSB & 0.9925 & 0.0122 & 0.0470 & 0.0771 & 0.9162 & 0.9635 & 91.4524 & 1.96 & 177.38 \\
&  & \macesb{} (ours) & \textbf{1.0000} & \textbf{0.0038} & \textbf{0.0248} & \textbf{0.0269} & \textbf{0.9724} & \textbf{0.9908} & \textbf{87.2876} & \underline{1.13} & 174.51 \\
\midrule
\multirow{7}{*}{Food-101}
& \multirow{4}{*}{Latent} & Stable Diffusion & 0.9791 & 0.3419 & 0.4199 & 0.8196 & 0.1701 & 0.2381 & 151.9260 & 31.96 & 156.24 \\
&  & LDCE & 0.9771 & 0.0442 & 0.0730 & 0.3204 & 0.6372 & 0.8135 & 36.8060 & 9.07 & 72.01 \\
&  & DiG-IN & 0.9890 & 0.0568 & 0.0909 & 0.2745 & 0.7238 & 0.6761 & 30.9616 & 1.08 & 66.18 \\
&  & \mace{} (ours) & \textbf{1.0000} & \underline{0.0051} & 0.0307 & \underline{0.0383} & \underline{0.9645} & \underline{0.9894} & \textbf{25.9439} & \underline{0.35} & \textbf{58.55} \\
& \multirow{3}{*}{Pixel} & MaskDiME & 0.8644 & 0.0097 & \textbf{0.0180} & 0.0908 & 0.9328 & 0.9843 & \textbf{25.9439} & 0.38 & \underline{58.71} \\
&  & RCSB & \textbf{1.0000} & 0.0124 & 0.0433 & 0.0827 & 0.9122 & 0.9716 & 26.9689 & \textbf{0.29} & 60.81 \\
&  & \macesb{} (ours) & \textbf{1.0000} & \textbf{0.0038} & \underline{0.0227} & \textbf{0.0316} & \textbf{0.9688} & \textbf{0.9915} & 26.1929 & 0.47 & 59.39 \\
\midrule
\multirow{7}{*}{Oxford Pets}
& \multirow{4}{*}{Latent} & Stable Diffusion & 0.7755 & 0.3196 & 0.3906 & 0.8445 & 0.1598 & 0.1316 & 140.7207 & 16.10 & 142.70 \\
&  & LDCE & 0.8878 & 0.0381 & 0.0678 & 0.2771 & 0.6579 & 0.7938 & 56.6523 & 15.59 & 87.07 \\
&  & DiG-IN & 0.9648 & 0.0500 & 0.0899 & 0.2497 & 0.7325 & 0.5175 & 57.9469 & 12.45 & 86.74 \\
&  & \mace{} (ours) & \textbf{1.0000} & \underline{0.0060} & 0.0363 & \underline{0.0352} & \underline{0.9618} & \underline{0.9821} & 35.5314 & \underline{3.63} & 61.68 \\
& \multirow{3}{*}{Pixel} & MaskDiME & 0.6901 & 0.0117 & \underline{0.0248} & 0.0797 & 0.9189 & 0.9805 & \textbf{32.7855} & \textbf{3.23} & \textbf{58.35} \\
&  & RCSB & 0.9891 & 0.0114 & 0.0435 & 0.0657 & 0.9207 & 0.9492 & 41.7721 & 7.80 & 70.35 \\
&  & \macesb{} (ours) & \underline{0.9992} & \textbf{0.0034} & \textbf{0.0225} & \textbf{0.0249} & \textbf{0.9732} & \textbf{0.9877} & \underline{34.0230} & 3.82 & \underline{60.11} \\
\midrule
\multirow{7}{*}{CUB-200}
& \multirow{4}{*}{Latent} & Stable Diffusion & 0.4044 & 0.2864 & 0.3487 & 0.8460 & 0.1632 & 0.2422 & 66.5195 & 16.05 & 68.89 \\
&  & LDCE & 0.9511 & 0.0325 & 0.0645 & 0.2216 & 0.7491 & 0.9128 & 28.3878 & 5.32 & 45.87 \\
&  & DiG-IN & 0.9636 & 0.0324 & 0.0670 & 0.1615 & 0.8392 & 0.9082 & 27.9030 & 6.60 & 44.99 \\
&  & \mace{} (ours) & \textbf{1.0000} & \underline{0.0053} & 0.0351 & \underline{0.0290} & \underline{0.9698} & \underline{0.9934} & \textbf{22.1957} & \underline{3.38} & \textbf{36.50} \\
& \multirow{3}{*}{Pixel} & MaskDiME & 0.8514 & 0.0090 & \textbf{0.0216} & 0.0669 & 0.9433 & 0.9894 & \underline{22.3795} & \textbf{3.27} & 38.54 \\
&  & RCSB & \textbf{1.0000} & 0.0121 & 0.0485 & 0.0733 & 0.9230 & 0.9632 & 32.3073 & 7.32 & 50.16 \\
&  & \macesb{} (ours) & 0.9994 & \textbf{0.0034} & \underline{0.0243} & \textbf{0.0245} & \textbf{0.9758} & \textbf{0.9937} & 23.4001 & 4.08 & \underline{38.45} \\
\bottomrule
\end{tabular}%
}
\caption{Quantitative comparison of counterfactual generation methods with the class ranked 6th by CLIP as the target. We report target-class validity, proximity and realism metrics. All metrics except validity are computed on the common valid set. KID is reported in units of $10^{-3}$. Bold and underlined values indicate the best and second-best results, respectively, within each dataset and metric column. Gen. indicates the diffusion model type (latent or pixel space).}
\label{tab:appendix_rank5}
\end{table*}
\begin{table*}[t]
    \centering
    \small
    \setlength{\tabcolsep}{3pt}
\resizebox{\textwidth}{!}{%
\begin{tabular}{lllcccccccccc}
\toprule
 & & & Validity & \multicolumn{5}{c}{Proximity} & \multicolumn{3}{c}{Realism} \\
\cmidrule(lr){4-4}\cmidrule(lr){5-9}\cmidrule(lr){10-12}
Dataset & Gen. & Method/Variant & Target $\uparrow$ &
$\ell_1^{\text{norm}}\downarrow$ & $\ell_2^{\text{norm}}\downarrow$ & LPIPS $\downarrow$ & SSIM $\uparrow$ & \sss{} $\uparrow$ &
FID $\downarrow$ & KID $\downarrow$ & sFID $\downarrow$ \\
\midrule
\multirow{7}{*}{ImageNet}
& \multirow{4}{*}{Latent} & Stable Diffusion & 0.6413 & 0.3274 & 0.3975 & 0.8473 & 0.1373 & 0.0510 & 128.4713 & 5.25 & \textbf{187.36} \\
&  & LDCE & 0.8726 & 0.0513 & 0.0917 & 0.3339 & 0.6044 & 0.7420 & 123.5433 & 16.96 & 202.98 \\
&  & DiG-IN & 0.8694 & 0.0604 & 0.1054 & 0.3186 & 0.6827 & 0.4229 & 114.6152 & 17.48 & \underline{190.55} \\
&  & \mace{} (ours) & \textbf{1.0000} & \underline{0.0064} & 0.0379 & \underline{0.0364} & \underline{0.9608} & \underline{0.9840} & 110.0030 & 1.23 & 191.98 \\
& \multirow{3}{*}{Pixel} & MaskDiME & 0.6392 & 0.0132 & \underline{0.0285} & 0.1033 & 0.9069 & 0.9780 & \textbf{108.4820} & \textbf{-0.57} & 191.81 \\
&  & RCSB & 0.9893 & 0.0119 & 0.0458 & 0.0765 & 0.9179 & 0.9576 & 112.2639 & 3.80 & 195.87 \\
&  & \macesb{} (ours) & \textbf{1.0000} & \textbf{0.0038} & \textbf{0.0248} & \textbf{0.0282} & \textbf{0.9716} & \textbf{0.9895} & \underline{109.5856} & \underline{0.57} & 191.77 \\
\midrule
\multirow{7}{*}{Food-101}
& \multirow{4}{*}{Latent} & Stable Diffusion & 0.9611 & 0.3410 & 0.4184 & 0.8199 & 0.1693 & 0.1882 & 155.9119 & 32.59 & 160.64 \\
&  & LDCE & 0.9751 & 0.0442 & 0.0729 & 0.3216 & 0.6390 & 0.8038 & 37.8476 & 7.80 & 73.59 \\
&  & DiG-IN & 0.9870 & 0.0585 & 0.0932 & 0.2867 & 0.7171 & 0.6343 & 32.9963 & 1.05 & 70.28 \\
&  & \mace{} (ours) & \textbf{1.0000} & \underline{0.0058} & 0.0333 & \underline{0.0441} & \underline{0.9604} & \underline{0.9869} & \underline{26.8574} & -0.10 & \textbf{61.07} \\
& \multirow{3}{*}{Pixel} & MaskDiME & 0.8495 & 0.0097 & \textbf{0.0181} & 0.0915 & 0.9326 & 0.9837 & \textbf{26.8093} & 0.14 & \underline{61.18} \\
&  & RCSB & 0.9960 & 0.0126 & 0.0436 & 0.0848 & 0.9111 & 0.9680 & 28.6863 & \textbf{-0.15} & 64.56 \\
&  & \macesb{} (ours) & \textbf{1.0000} & \textbf{0.0040} & \underline{0.0236} & \textbf{0.0344} & \textbf{0.9666} & \textbf{0.9899} & 27.0198 & \underline{-0.11} & 62.20 \\
\midrule
\multirow{7}{*}{Oxford Pets}
& \multirow{4}{*}{Latent} & Stable Diffusion & 0.7286 & 0.3191 & 0.3902 & 0.8405 & 0.1591 & 0.0908 & 141.0632 & 18.00 & 145.00 \\
&  & LDCE & 0.8308 & 0.0379 & 0.0678 & 0.2823 & 0.6574 & 0.7645 & 73.4873 & 23.82 & 110.78 \\
&  & DiG-IN & 0.9497 & 0.0502 & 0.0910 & 0.2506 & 0.7297 & 0.4738 & 70.8204 & 14.54 & 100.98 \\
&  & \mace{} (ours) & \textbf{0.9992} & \underline{0.0068} & 0.0394 & \underline{0.0404} & \underline{0.9570} & 0.9736 & 47.2019 & 5.48 & 77.37 \\
& \multirow{3}{*}{Pixel} & MaskDiME & 0.6181 & 0.0120 & \underline{0.0254} & 0.0821 & 0.9172 & \underline{0.9775} & \textbf{41.0133} & \textbf{2.88} & \textbf{71.48} \\
&  & RCSB & 0.9791 & 0.0117 & 0.0443 & 0.0678 & 0.9189 & 0.9375 & 54.2109 & 9.77 & 86.94 \\
&  & \macesb{} (ours) & \textbf{0.9992} & \textbf{0.0036} & \textbf{0.0233} & \textbf{0.0270} & \textbf{0.9716} & \textbf{0.9847} & \underline{42.9389} & \underline{4.89} & \underline{73.62} \\
\midrule
\multirow{7}{*}{CUB-200}
& \multirow{4}{*}{Latent} & Stable Diffusion & 0.3636 & 0.2810 & 0.3424 & 0.8491 & 0.1686 & 0.2512 & 67.1764 & 14.03 & 75.25 \\
&  & LDCE & 0.9129 & 0.0323 & 0.0635 & 0.2218 & 0.7475 & 0.9118 & 31.6990 & 2.57 & 53.98 \\
&  & DiG-IN & 0.9335 & 0.0329 & 0.0675 & 0.1663 & 0.8383 & 0.8910 & 30.5943 & 4.74 & 50.35 \\
&  & \mace{} (ours) & \underline{0.9994} & \underline{0.0054} & 0.0357 & \underline{0.0307} & \underline{0.9688} & \textbf{0.9936} & \textbf{23.7325} & \textbf{1.24} & \textbf{43.38} \\
& \multirow{3}{*}{Pixel} & MaskDiME & 0.8169 & 0.0088 & \textbf{0.0211} & 0.0672 & 0.9446 & 0.9881 & \underline{24.7363} & \underline{1.63} & \underline{45.39} \\
&  & RCSB & \textbf{1.0000} & 0.0122 & 0.0482 & 0.0743 & 0.9224 & 0.9596 & 34.8851 & 3.35 & 57.29 \\
&  & \macesb{} (ours) & \underline{0.9994} & \textbf{0.0033} & \underline{0.0239} & \textbf{0.0244} & \textbf{0.9763} & \underline{0.9930} & 25.4425 & 1.93 & 45.52 \\
\bottomrule
\end{tabular}%
}
\caption{Quantitative comparison of counterfactual generation methods with the class ranked 11th by CLIP as the target. We report target-class validity, proximity and realism metrics. All metrics except validity are computed on the common valid set. KID is reported in units of $10^{-3}$. Bold and underlined values indicate the best and second-best results, respectively, within each dataset and metric column. Gen. indicates the diffusion model type (latent or pixel space).}
\label{tab:appendix_rank10}
\end{table*}

\begin{table*}[t]
    \centering
    \small
    \setlength{\tabcolsep}{3pt}
\resizebox{\textwidth}{!}{%
\begin{tabular}{lllccccccccc}
\toprule
 & & & Validity & \multicolumn{5}{c}{Proximity} & \multicolumn{3}{c}{Realism} \\
\cmidrule(lr){4-4}\cmidrule(lr){5-9}\cmidrule(lr){10-12}
Dataset & Gen. & Method/Variant & Target $\uparrow$ &
$\ell_1^{\text{norm}}\downarrow$ & $\ell_2^{\text{norm}}\downarrow$ & LPIPS $\downarrow$ & SSIM $\uparrow$ & \sss{} $\uparrow$ &
FID $\downarrow$ & KID $\downarrow$ & sFID $\downarrow$ \\
\midrule
\multirow{6}{*}{ImageNet}
& \multirow{3}{*}{Latent} & Stable Diffusion & 0.7302 & 0.3393 & 0.4102 & 0.8478 & 0.1228 & 0.0020 & \underline{142.4853} & \textbf{2.52} & \textbf{209.60} \\
&  & LDCE & 0.6617 & 0.0526 & 0.0922 & 0.3502 & 0.5892 & 0.6010 & 176.9603 & 49.14 & 265.05 \\
&  & \mace{} (ours) & \textbf{0.9989} & \underline{0.0106} & 0.0508 & \underline{0.0616} & \underline{0.9366} & 0.9522 & 147.0146 & 11.75 & 232.50 \\
& \multirow{3}{*}{Pixel} & MaskDiME & 0.4561 & 0.0138 & \textbf{0.0294} & 0.1129 & 0.9029 & \textbf{0.9730} & \textbf{141.2997} & \underline{5.24} & \underline{229.41} \\
&  & RCSB & 0.9026 & 0.0125 & 0.0473 & 0.0835 & 0.9138 & 0.9368 & 147.8486 & 12.11 & 236.16 \\
&  & \macesb{} (ours) & \underline{0.9957} & \textbf{0.0061} & \underline{0.0313} & \textbf{0.0434} & \textbf{0.9558} & \underline{0.9719} & 143.8117 & 10.13 & 230.29 \\
\midrule
\multirow{6}{*}{Food-101}
& \multirow{3}{*}{Latent} & Stable Diffusion & 0.9581 & 0.3411 & 0.4184 & 0.8209 & 0.1668 & 0.1303 & 144.7271 & 33.24 & 153.00 \\
&  & LDCE & 0.9103 & 0.0445 & 0.0735 & 0.3238 & 0.6355 & 0.7928 & 42.2058 & 10.62 & 78.39 \\
&  & \mace{} (ours) & \textbf{1.0000} & \underline{0.0072} & 0.0383 & \underline{0.0545} & \underline{0.9521} & 0.9830 & \underline{28.9511} & 0.97 & \underline{63.91} \\
& \multirow{3}{*}{Pixel} & MaskDiME & 0.7876 & 0.0100 & \textbf{0.0186} & 0.0935 & 0.9312 & \underline{0.9845} & \textbf{28.4608} & \textbf{0.42} & \textbf{63.84} \\
&  & RCSB & 0.9900 & 0.0128 & 0.0444 & 0.0874 & 0.9093 & 0.9630 & 31.4070 & 1.46 & 67.63 \\
&  & \macesb{} (ours) & \underline{0.9980} & \textbf{0.0047} & \underline{0.0256} & \textbf{0.0393} & \textbf{0.9612} & \textbf{0.9873} & 29.1514 & \underline{0.58} & 64.68 \\
\midrule
\multirow{6}{*}{Oxford Pets}
& \multirow{3}{*}{Latent} & Stable Diffusion & 0.8576 & 0.3191 & 0.3894 & 0.8424 & 0.1579 & 0.0258 & 134.1449 & 13.87 & 143.45 \\
&  & LDCE & 0.8032 & 0.0378 & 0.0673 & 0.2827 & 0.6601 & 0.7076 & 78.5663 & 32.86 & 120.58 \\
&  & \mace{} (ours) & \textbf{1.0000} & \underline{0.0080} & 0.0434 & \underline{0.0471} & \underline{0.9504} & 0.9595 & 45.7159 & 8.96 & 77.84 \\
& \multirow{3}{*}{Pixel} & MaskDiME & 0.5871 & 0.0120 & \underline{0.0254} & 0.0818 & 0.9182 & \underline{0.9763} & \textbf{36.6712} & \textbf{1.97} & \textbf{70.33} \\
&  & RCSB & 0.9648 & 0.0118 & 0.0447 & 0.0703 & 0.9179 & 0.9312 & 54.3649 & 13.62 & 90.93 \\
&  & \macesb{} (ours) & \underline{0.9992} & \textbf{0.0041} & \textbf{0.0250} & \textbf{0.0304} & \textbf{0.9675} & \textbf{0.9826} & \underline{39.2944} & \underline{4.72} & \underline{72.45} \\
\midrule
\multirow{6}{*}{CUB-200}
& \multirow{3}{*}{Latent} & Stable Diffusion & 0.5273 & 0.2869 & 0.3483 & 0.8506 & 0.1669 & 0.1617 & 62.1600 & 12.27 & 67.74 \\
&  & LDCE & 0.7618 & 0.0337 & 0.0655 & 0.2305 & 0.7408 & 0.8988 & 29.3068 & 3.89 & 51.13 \\
&  & \mace{} (ours) & \textbf{1.0000} & \underline{0.0080} & 0.0456 & \underline{0.0450} & \underline{0.9559} & 0.9835 & 21.9815 & 3.23 & \textbf{38.20} \\
& \multirow{3}{*}{Pixel} & MaskDiME & 0.8163 & 0.0092 & \textbf{0.0220} & 0.0699 & 0.9430 & \underline{0.9876} & \textbf{20.5704} & \textbf{2.11} & \underline{38.50} \\
&  & RCSB & 0.9918 & 0.0127 & 0.0495 & 0.0793 & 0.9188 & 0.9527 & 34.8213 & 7.02 & 56.87 \\
&  & \macesb{} (ours) & \underline{0.9975} & \textbf{0.0043} & \underline{0.0271} & \textbf{0.0296} & \textbf{0.9696} & \textbf{0.9894} & \underline{21.4479} & \underline{3.01} & 39.16 \\
\bottomrule
\end{tabular}%
}
\caption{Quantitative comparison with the class at about 50\% of CLIP's ranking as the target (target rank/number of classes: ImageNet 500/1000, Food-101 51/101, Oxford Pets 19/37, CUB-200 100/200), on the full benchmark of each dataset. Validity is computed over all instances; all other metrics on the common valid set (ImageNet 230, Food-101 713, Oxford Pets 552, CUB-200 584 instances). FID and sFID depend on the number of instances and are only comparable within a dataset. KID is reported in units of $10^{-3}$. Bold and underlined values indicate the best and second-best results, respectively, within each dataset and metric column. Gen. indicates the diffusion model type (latent or pixel space).}
\label{tab:appendix_rank_low50}
\end{table*}
\begin{table*}[t]
    \centering
    \small
    \setlength{\tabcolsep}{3pt}
\resizebox{\textwidth}{!}{%
\begin{tabular}{lllccccccccc}
\toprule
 & & & Validity & \multicolumn{5}{c}{Proximity} & \multicolumn{3}{c}{Realism} \\
\cmidrule(lr){4-4}\cmidrule(lr){5-9}\cmidrule(lr){10-12}
Dataset & Gen. & Method/Variant & Target $\uparrow$ &
$\ell_1^{\text{norm}}\downarrow$ & $\ell_2^{\text{norm}}\downarrow$ & LPIPS $\downarrow$ & SSIM $\uparrow$ & \sss{} $\uparrow$ &
FID $\downarrow$ & KID $\downarrow$ & sFID $\downarrow$ \\
\midrule
\multirow{6}{*}{ImageNet}
& \multirow{3}{*}{Latent} & Stable Diffusion & 0.7645 & 0.3345 & 0.4058 & 0.8565 & 0.1226 & -0.0008 & \underline{140.1876} & \underline{4.66} & \textbf{206.58} \\
&  & LDCE & 0.6606 & 0.0511 & 0.0895 & 0.3537 & 0.5959 & 0.5776 & 185.2450 & 61.80 & 267.53 \\
&  & \mace{} (ours) & \textbf{0.9989} & \underline{0.0113} & 0.0521 & \underline{0.0673} & \underline{0.9312} & 0.9548 & 147.7722 & 11.37 & 229.09 \\
& \multirow{3}{*}{Pixel} & MaskDiME & 0.4529 & 0.0137 & \textbf{0.0291} & 0.1163 & 0.9013 & \textbf{0.9735} & \textbf{139.0652} & \textbf{3.04} & \underline{223.34} \\
&  & RCSB & 0.8887 & 0.0126 & 0.0473 & 0.0849 & 0.9129 & 0.9533 & 149.3598 & 10.21 & 232.38 \\
&  & \macesb{} (ours) & \underline{0.9979} & \textbf{0.0065} & \underline{0.0323} & \textbf{0.0478} & \textbf{0.9524} & \underline{0.9689} & 141.5949 & 5.65 & 224.60 \\
\midrule
\multirow{6}{*}{Food-101}
& \multirow{3}{*}{Latent} & Stable Diffusion & 0.9511 & 0.3427 & 0.4200 & 0.8225 & 0.1662 & 0.1235 & 147.2463 & 29.46 & 155.28 \\
&  & LDCE & 0.9013 & 0.0446 & 0.0733 & 0.3275 & 0.6325 & 0.7743 & 45.0927 & 12.08 & 81.51 \\
&  & \mace{} (ours) & \textbf{1.0000} & \underline{0.0081} & 0.0411 & \underline{0.0604} & \underline{0.9466} & 0.9804 & 32.5014 & 2.63 & \underline{65.72} \\
& \multirow{3}{*}{Pixel} & MaskDiME & 0.7777 & 0.0101 & \textbf{0.0187} & 0.0952 & 0.9305 & \underline{0.9842} & \textbf{31.1018} & \textbf{1.70} & \textbf{64.63} \\
&  & RCSB & \underline{0.9840} & 0.0129 & 0.0446 & 0.0887 & 0.9083 & 0.9609 & 34.7945 & 2.97 & 69.81 \\
&  & \macesb{} (ours) & \textbf{1.0000} & \textbf{0.0052} & \underline{0.0271} & \textbf{0.0431} & \textbf{0.9575} & \textbf{0.9865} & \underline{32.4626} & \underline{2.39} & 66.55 \\
\midrule
\multirow{6}{*}{Oxford Pets}
& \multirow{3}{*}{Latent} & Stable Diffusion & 0.8652 & 0.3219 & 0.3931 & 0.8440 & 0.1602 & 0.0180 & 131.6682 & 15.19 & 137.81 \\
&  & LDCE & 0.8040 & 0.0383 & 0.0684 & 0.2855 & 0.6569 & 0.7086 & 79.6270 & 27.81 & 125.45 \\
&  & \mace{} (ours) & \textbf{1.0000} & \underline{0.0085} & 0.0453 & \underline{0.0491} & \underline{0.9491} & 0.9523 & 49.0061 & 6.86 & 82.08 \\
& \multirow{3}{*}{Pixel} & MaskDiME & 0.5921 & 0.0121 & \underline{0.0257} & 0.0835 & 0.9174 & \underline{0.9764} & \textbf{37.1176} & \textbf{0.78} & \textbf{69.76} \\
&  & RCSB & 0.9456 & 0.0119 & 0.0451 & 0.0706 & 0.9173 & 0.9267 & 55.0960 & 8.72 & 92.07 \\
&  & \macesb{} (ours) & \underline{0.9983} & \textbf{0.0042} & \textbf{0.0254} & \textbf{0.0307} & \textbf{0.9674} & \textbf{0.9780} & \underline{40.4559} & \underline{2.00} & \underline{72.81} \\
\midrule
\multirow{6}{*}{CUB-200}
& \multirow{3}{*}{Latent} & Stable Diffusion & 0.5843 & 0.2879 & 0.3500 & 0.8563 & 0.1661 & 0.1615 & 64.0436 & 11.92 & 71.54 \\
&  & LDCE & 0.7066 & 0.0340 & 0.0664 & 0.2316 & 0.7385 & 0.8782 & 32.2046 & 4.78 & 55.62 \\
&  & \mace{} (ours) & \textbf{0.9994} & \underline{0.0093} & 0.0503 & \underline{0.0524} & \underline{0.9508} & 0.9790 & 24.1825 & 3.85 & 40.58 \\
& \multirow{3}{*}{Pixel} & MaskDiME & 0.8006 & 0.0094 & \textbf{0.0226} & 0.0696 & 0.9426 & \textbf{0.9853} & \textbf{20.8363} & \underline{2.28} & \textbf{38.71} \\
&  & RCSB & \underline{0.9956} & 0.0128 & 0.0498 & 0.0796 & 0.9188 & 0.9485 & 34.1078 & 3.67 & 57.79 \\
&  & \macesb{} (ours) & \textbf{0.9994} & \textbf{0.0048} & \underline{0.0289} & \textbf{0.0323} & \textbf{0.9672} & \underline{0.9851} & \underline{22.4510} & \textbf{2.11} & \underline{40.19} \\
\bottomrule
\end{tabular}%
}
\caption{Quantitative comparison with the class at about 70\% of CLIP's ranking as the target (target rank/number of classes: ImageNet 700/1000, Food-101 71/101, Oxford Pets 26/37, CUB-200 140/200), on the full benchmark of each dataset. Validity is computed over all instances; all other metrics on the common valid set (ImageNet 224, Food-101 681, Oxford Pets 541, CUB-200 619 instances). FID and sFID depend on the number of instances and are only comparable within a dataset. KID is reported in units of $10^{-3}$. Bold and underlined values indicate the best and second-best results, respectively, within each dataset and metric column. Gen. indicates the diffusion model type (latent or pixel space).}
\label{tab:appendix_rank_low70}
\end{table*}

\begin{table*}[tp]
    \centering
    \small
    \setlength{\tabcolsep}{4pt}
\begin{tabular}{lccccccc}
\toprule
Text prompt & Validity $\uparrow$ & $\ell_1^{\text{norm}}\downarrow$ & LPIPS $\downarrow$ & SSIM $\uparrow$ & $r=0.05$ (\%) & Generations & CMV \\
\midrule
Target class name (default) & 0.999 & 0.0047 & 0.028 & 0.969 & 88.1 & 1.98 & 0.104 \\
Empty prompt & 1.000 & 0.0046 & 0.027 & 0.970 & 91.2 & 1.81 & 0.094 \\
\bottomrule
\end{tabular}
\caption{\mace{} with the target class name or an empty prompt as the text condition of Stable Diffusion, with CLIP guidance, mask construction, and all other settings unchanged (full benchmark, mean over the four datasets). Validity, $r=0.05$, Generations, and CMV are computed over all instances, proximity on the instances valid for both variants. $r=0.05$: share of instances for which the smallest mask suffices; Generations: mean number of generations of the adaptive search; CMV: mean target top-1 rate of the five held-out CLIPs.}
\label{tab:appendix_no_prompt}
\end{table*}

\begin{table*}[t]
    \centering
    \small
    \setlength{\tabcolsep}{3pt}
    \renewcommand{\arraystretch}{0.9}
\resizebox{\textwidth}{!}{
\begin{tabular}{lllcccccccccc}
\toprule
 & & & Validity & \multicolumn{5}{c}{Proximity} & \multicolumn{3}{c}{Realism} & \\
\cmidrule(lr){5-9}\cmidrule(lr){10-12}
Dataset & Gen. & Method/Variant & Target $\uparrow$ & $\ell_1^{\text{norm}}\downarrow$ & $\ell_2^{\text{norm}}\downarrow$ & LPIPS $\downarrow$ & SSIM $\uparrow$ & \sss{} $\uparrow$ & FID $\downarrow$ & KID $\downarrow$ & sFID $\downarrow$ & CMV \\
\midrule
\multirow{7}{*}{ImageNet} & \multirow{4}{*}{Latent} & Stable Diffusion & 0.7686 & 0.3332 & 0.4043 & 0.8315 & 0.1293 & 0.2055 & 117.25 & 4.45 & \textbf{179.75} & 0.7240 \\
 &  & LDCE & 0.9131 & 0.0537 & 0.0949 & 0.3265 & 0.5951 & 0.8362 & 112.90 & 8.81 & 192.67 & 0.2716 \\
 &  & DiG-IN & 0.9131 & 0.0560 & 0.0981 & 0.2825 & 0.6987 & 0.6473 & 106.28 & 8.77 & \underline{182.24} & 0.6383 \\
 &  & \mace{} (ours) & \textbf{0.9990} & \underline{0.0046} & 0.0297 & \underline{0.0230} & \underline{0.9705} & \textbf{0.9945} & \textbf{102.74} & \underline{1.05} & 184.70 & 0.1169 \\
 & \multirow{3}{*}{Pixel} & MaskDiME & 0.4712 & 0.0144 & \underline{0.0290} & 0.1029 & 0.9078 & 0.9851 & 104.21 & \textbf{0.61} & 185.93 & 0.0921 \\
 &  & RCSB & 0.9560 & 0.0101 & 0.0407 & 0.0635 & 0.9279 & 0.9776 & 106.81 & 2.94 & 190.52 & 0.1527 \\
 &  & \macesb{} (ours) & \underline{0.9979} & \textbf{0.0032} & \textbf{0.0222} & \textbf{0.0218} & \textbf{0.9762} & \underline{0.9943} & \underline{103.93} & 1.20 & 185.45 & 0.1160 \\
\midrule
\multirow{7}{*}{Food-101} & \multirow{4}{*}{Latent} & Stable Diffusion & 0.9613 & 0.3362 & 0.4134 & 0.8214 & 0.1786 & 0.2956 & 156.22 & 34.19 & 164.51 & 0.9231 \\
 &  & LDCE & 0.9652 & 0.0421 & 0.0696 & 0.3101 & 0.6494 & 0.8364 & 45.18 & 6.27 & 82.76 & 0.4056 \\
 &  & DiG-IN & 0.9921 & 0.0503 & 0.0807 & 0.2441 & 0.7453 & 0.7341 & 40.48 & 1.37 & 79.13 & 0.9072 \\
 &  & \mace{} (ours) & \textbf{1.0000} & \underline{0.0039} & 0.0247 & \underline{0.0276} & \underline{0.9709} & \underline{0.9936} & \underline{36.80} & 0.64 & \textbf{69.76} & 0.1080 \\
 & \multirow{3}{*}{Pixel} & MaskDiME & 0.5501 & 0.0093 & \textbf{0.0162} & 0.0854 & 0.9385 & 0.9916 & \textbf{36.75} & 0.12 & \underline{70.45} & 0.0691 \\
 &  & RCSB & 0.9821 & 0.0099 & 0.0365 & 0.0693 & 0.9274 & 0.9804 & 37.99 & \underline{-0.23} & 72.93 & 0.1635 \\
 &  & \macesb{} (ours) & \textbf{1.0000} & \textbf{0.0032} & \underline{0.0201} & \textbf{0.0264} & \textbf{0.9725} & \textbf{0.9945} & 37.11 & \textbf{-0.24} & 70.77 & 0.1202 \\
\midrule
\multirow{7}{*}{Oxford Pets} & \multirow{4}{*}{Latent} & Stable Diffusion & 0.8714 & 0.3115 & 0.3813 & 0.8359 & 0.1490 & 0.2852 & 134.65 & 23.97 & 143.17 & 0.8954 \\
 &  & LDCE & 0.9230 & 0.0410 & 0.0714 & 0.2887 & 0.6332 & 0.8526 & 59.02 & 10.92 & 87.59 & 0.3032 \\
 &  & DiG-IN & 0.9897 & 0.0461 & 0.0811 & 0.2185 & 0.7404 & 0.7576 & 55.32 & 8.34 & 87.39 & 0.8700 \\
 &  & \mace{} (ours) & \textbf{1.0000} & \underline{0.0041} & 0.0276 & \underline{0.0219} & \underline{0.9721} & \underline{0.9907} & \underline{42.33} & 4.07 & 70.45 & 0.1087 \\
 & \multirow{3}{*}{Pixel} & MaskDiME & 0.4286 & 0.0142 & \underline{0.0272} & 0.0924 & 0.9127 & 0.9862 & \textbf{41.09} & \textbf{2.56} & \textbf{70.13} & 0.0771 \\
 &  & RCSB & 0.9778 & 0.0099 & 0.0391 & 0.0539 & 0.9281 & 0.9732 & 48.87 & 5.89 & 78.09 & 0.2073 \\
 &  & \macesb{} (ours) & \textbf{1.0000} & \textbf{0.0027} & \textbf{0.0189} & \textbf{0.0189} & \textbf{0.9781} & \textbf{0.9912} & 42.55 & \underline{3.47} & \underline{70.39} & 0.1152 \\
\midrule
\multirow{7}{*}{CUB-200} & \multirow{4}{*}{Latent} & Stable Diffusion & 0.6096 & 0.2848 & 0.3459 & 0.8411 & 0.1708 & 0.2572 & 63.26 & 13.89 & 69.07 & 0.4350 \\
 &  & LDCE & 0.9585 & 0.0338 & 0.0650 & 0.2270 & 0.7345 & 0.9113 & 26.87 & 3.47 & 45.97 & 0.2420 \\
 &  & DiG-IN & 0.9895 & 0.0299 & 0.0599 & 0.1430 & 0.8464 & 0.9227 & 25.55 & 3.93 & 41.63 & 0.5371 \\
 &  & \mace{} (ours) & \textbf{1.0000} & \underline{0.0044} & 0.0307 & \underline{0.0215} & \underline{0.9731} & \textbf{0.9952} & \textbf{18.69} & \textbf{0.44} & \textbf{35.63} & 0.1560 \\
 & \multirow{3}{*}{Pixel} & MaskDiME & 0.5874 & 0.0086 & \textbf{0.0183} & 0.0618 & 0.9479 & 0.9914 & \underline{19.01} & \underline{0.94} & \underline{36.93} & 0.0911 \\
 &  & RCSB & \underline{0.9994} & 0.0096 & 0.0411 & 0.0580 & 0.9363 & 0.9746 & 27.96 & 3.13 & 46.55 & 0.2396 \\
 &  & \macesb{} (ours) & \underline{0.9994} & \textbf{0.0032} & \underline{0.0229} & \textbf{0.0203} & \textbf{0.9779} & \underline{0.9948} & 20.05 & 1.74 & 37.59 & 0.1544 \\
\bottomrule
\end{tabular}}
\caption{Quantitative comparison of counterfactual generation methods with OpenAI CLIP ViT-L/14 as the explained model on the full benchmark (955/1007/1260/1711 instances for ImageNet/Food-101/Oxford Pets/CUB-200). We report target-class validity, proximity and realism metrics and cross-model validity (CMV) over five held-out CLIP models. All metrics except validity and CMV are computed on the common valid set (338/526/467/612 instances for ImageNet/Food-101/Oxford Pets/CUB-200). KID is reported in units of $10^{-3}$. Bold and underlined values indicate the best and second-best results, respectively, within each dataset and metric column. Gen. indicates the diffusion model type (latent or pixel space).}
\label{tab:appendix_target_l14}
\end{table*}

\begin{table*}[t]
    \centering
    \small
    \setlength{\tabcolsep}{3pt}
    \renewcommand{\arraystretch}{0.9}
\resizebox{\textwidth}{!}{
\begin{tabular}{lllcccccccccc}
\toprule
 & & & Validity & \multicolumn{5}{c}{Proximity} & \multicolumn{3}{c}{Realism} & \\
\cmidrule(lr){5-9}\cmidrule(lr){10-12}
Dataset & Gen. & Method/Variant & Target $\uparrow$ & $\ell_1^{\text{norm}}\downarrow$ & $\ell_2^{\text{norm}}\downarrow$ & LPIPS $\downarrow$ & SSIM $\uparrow$ & \sss{} $\uparrow$ & FID $\downarrow$ & KID $\downarrow$ & sFID $\downarrow$ & CMV \\
\midrule
\multirow{7}{*}{ImageNet} & \multirow{4}{*}{Latent} & Stable Diffusion & 0.7138 & 0.3373 & 0.4091 & 0.8335 & 0.1301 & 0.2102 & 101.23 & 4.20 & \underline{159.26} & 0.7700 \\
 &  & LDCE & 0.9838 & 0.0530 & 0.0947 & 0.3286 & 0.6011 & 0.8278 & 84.27 & 6.35 & 166.03 & 0.3961 \\
 &  & DiG-IN & 0.9590 & 0.0536 & 0.0960 & 0.2669 & 0.7101 & 0.6598 & 79.34 & 4.88 & \textbf{157.30} & 0.7257 \\
 &  & \mace{} (ours) & \textbf{0.9989} & \underline{0.0045} & 0.0320 & \textbf{0.0225} & \underline{0.9736} & \textbf{0.9944} & \textbf{75.92} & 0.10 & 162.39 & 0.0987 \\
 & \multirow{3}{*}{Pixel} & MaskDiME & 0.7559 & 0.0122 & \underline{0.0289} & 0.1101 & 0.9188 & 0.9843 & \underline{76.85} & \underline{0.01} & 163.77 & 0.0786 \\
 &  & RCSB & \textbf{0.9989} & 0.0115 & 0.0452 & 0.0738 & 0.9208 & 0.9805 & 77.22 & 0.58 & 162.87 & 0.1510 \\
 &  & \macesb{} (ours) & \textbf{0.9989} & \textbf{0.0032} & \textbf{0.0235} & \underline{0.0233} & \textbf{0.9773} & \underline{0.9934} & 76.92 & \textbf{-0.01} & 163.17 & 0.0983 \\
\midrule
\multirow{7}{*}{Food-101} & \multirow{4}{*}{Latent} & Stable Diffusion & 0.9235 & 0.3386 & 0.4164 & 0.8172 & 0.1754 & 0.3013 & 151.99 & 32.28 & 155.41 & 0.9251 \\
 &  & LDCE & 0.9889 & 0.0448 & 0.0743 & 0.3203 & 0.6356 & 0.8281 & 36.07 & 7.42 & 69.80 & 0.5394 \\
 &  & DiG-IN & 0.9960 & 0.0523 & 0.0843 & 0.2465 & 0.7394 & 0.7521 & 31.21 & 1.52 & 66.01 & 0.9376 \\
 &  & \mace{} (ours) & \textbf{1.0000} & \underline{0.0041} & 0.0289 & \underline{0.0276} & \underline{0.9741} & \textbf{0.9935} & \underline{27.64} & 1.07 & \textbf{59.54} & 0.0465 \\
 & \multirow{3}{*}{Pixel} & MaskDiME & 0.8469 & 0.0090 & \textbf{0.0189} & 0.1042 & 0.9392 & 0.9873 & 27.95 & 1.22 & 60.21 & 0.0340 \\
 &  & RCSB & \textbf{1.0000} & 0.0116 & 0.0415 & 0.0808 & 0.9164 & 0.9782 & 28.48 & \underline{0.90} & 61.05 & 0.1037 \\
 &  & \macesb{} (ours) & 0.9990 & \textbf{0.0032} & \underline{0.0224} & \textbf{0.0264} & \textbf{0.9748} & \textbf{0.9935} & \textbf{27.53} & \textbf{0.30} & \underline{59.99} & 0.0495 \\
\midrule
\multirow{7}{*}{Oxford Pets} & \multirow{4}{*}{Latent} & Stable Diffusion & 0.8024 & 0.3181 & 0.3898 & 0.8326 & 0.1594 & 0.3092 & 131.55 & 17.79 & 138.57 & 0.8896 \\
 &  & LDCE & 0.9864 & 0.0389 & 0.0690 & 0.2768 & 0.6532 & 0.8389 & 47.42 & 10.47 & 75.73 & 0.4178 \\
 &  & DiG-IN & 0.9949 & 0.0465 & 0.0835 & 0.2232 & 0.7495 & 0.7055 & 46.85 & 8.19 & 75.99 & 0.9388 \\
 &  & \mace{} (ours) & \textbf{1.0000} & \underline{0.0046} & 0.0323 & \underline{0.0246} & \underline{0.9718} & \underline{0.9905} & \underline{33.61} & \textbf{2.44} & \underline{58.47} & 0.0673 \\
 & \multirow{3}{*}{Pixel} & MaskDiME & 0.6802 & 0.0109 & \underline{0.0248} & 0.0952 & 0.9269 & 0.9852 & \textbf{32.57} & \underline{2.88} & \textbf{58.34} & 0.0361 \\
 &  & RCSB & \textbf{1.0000} & 0.0109 & 0.0416 & 0.0617 & 0.9223 & 0.9662 & 39.12 & 5.22 & 63.53 & 0.1959 \\
 &  & \macesb{} (ours) & \textbf{1.0000} & \textbf{0.0033} & \textbf{0.0231} & \textbf{0.0237} & \textbf{0.9756} & \textbf{0.9906} & 34.48 & 4.39 & 58.98 & 0.0716 \\
\midrule
\multirow{7}{*}{CUB-200} & \multirow{4}{*}{Latent} & Stable Diffusion & 0.3860 & 0.2861 & 0.3479 & 0.8360 & 0.1593 & 0.2812 & 68.24 & 11.85 & 72.89 & 0.4657 \\
 &  & LDCE & 0.9838 & 0.0343 & 0.0667 & 0.2278 & 0.7369 & 0.9179 & 28.48 & 3.98 & 48.35 & 0.2869 \\
 &  & DiG-IN & 0.9925 & 0.0313 & 0.0640 & 0.1482 & 0.8442 & 0.9165 & 28.63 & 4.31 & 46.55 & 0.6270 \\
 &  & \mace{} (ours) & \textbf{1.0000} & \underline{0.0043} & 0.0319 & \textbf{0.0215} & \underline{0.9759} & \textbf{0.9954} & \textbf{21.07} & \textbf{1.74} & \textbf{38.86} & 0.1183 \\
 & \multirow{3}{*}{Pixel} & MaskDiME & 0.8345 & 0.0082 & \textbf{0.0213} & 0.0767 & 0.9510 & 0.9904 & \underline{21.90} & \underline{1.82} & \underline{40.15} & 0.0735 \\
 &  & RCSB & \textbf{1.0000} & 0.0112 & 0.0454 & 0.0680 & 0.9274 & 0.9779 & 27.02 & 3.24 & 45.60 & 0.2261 \\
 &  & \macesb{} (ours) & \textbf{1.0000} & \textbf{0.0031} & \underline{0.0240} & \underline{0.0221} & \textbf{0.9786} & \underline{0.9949} & 22.53 & 2.05 & 40.65 & 0.1146 \\
\bottomrule
\end{tabular}}
\caption{Quantitative comparison of counterfactual generation methods with OpenAI CLIP ViT-B/32 as the explained model on the full benchmark (926/993/1179/1601 instances for ImageNet/Food-101/Oxford Pets/CUB-200). We report target-class validity, proximity and realism metrics and cross-model validity (CMV) over five held-out CLIP models. All metrics except validity and CMV are computed on the common valid set (506/772/669/522 instances for ImageNet/Food-101/Oxford Pets/CUB-200). KID is reported in units of $10^{-3}$. Bold and underlined values indicate the best and second-best results, respectively, within each dataset and metric column. Gen. indicates the diffusion model type (latent or pixel space).}
\label{tab:appendix_target_b32}
\end{table*}

\begin{table*}[t]
    \centering
    \small
    \setlength{\tabcolsep}{3pt}
    \renewcommand{\arraystretch}{0.9}
\resizebox{\textwidth}{!}{
\begin{tabular}{lllcccccccccc}
\toprule
 & & & Validity & \multicolumn{5}{c}{Proximity} & \multicolumn{3}{c}{Realism} & \\
\cmidrule(lr){5-9}\cmidrule(lr){10-12}
Dataset & Gen. & Method/Variant & Target $\uparrow$ & $\ell_1^{\text{norm}}\downarrow$ & $\ell_2^{\text{norm}}\downarrow$ & LPIPS $\downarrow$ & SSIM $\uparrow$ & \sss{} $\uparrow$ & FID $\downarrow$ & KID $\downarrow$ & sFID $\downarrow$ & CMV \\
\midrule
\multirow{7}{*}{ImageNet} & \multirow{4}{*}{Latent} & Stable Diffusion & 0.7202 & 0.3322 & 0.4042 & 0.8321 & 0.1289 & 0.1981 & 107.29 & 4.63 & \underline{163.48} & 0.6926 \\
 &  & LDCE & 0.9362 & 0.0520 & 0.0928 & 0.3247 & 0.6027 & 0.8205 & 90.21 & 6.86 & 170.50 & 0.3463 \\
 &  & DiG-IN & 0.9681 & 0.0530 & 0.0943 & 0.2654 & 0.7122 & 0.6742 & 85.35 & 5.19 & \textbf{163.46} & 0.6780 \\
 &  & \mace{} (ours) & 0.9290 & \underline{0.0124} & 0.0474 & \underline{0.0620} & \underline{0.9235} & \underline{0.9818} & \underline{81.99} & 0.59 & 165.68 & 0.1759 \\
 & \multirow{3}{*}{Pixel} & MaskDiME & 0.7366 & 0.0139 & \underline{0.0292} & 0.1069 & 0.9062 & 0.9798 & 82.41 & \underline{-0.12} & 167.00 & 0.1551 \\
 &  & RCSB & \underline{0.9835} & 0.0125 & 0.0484 & 0.0779 & 0.9168 & 0.9686 & 84.53 & 2.04 & 168.36 & 0.2140 \\
 &  & \macesb{} (ours) & \textbf{0.9990} & \textbf{0.0032} & \textbf{0.0218} & \textbf{0.0249} & \textbf{0.9753} & \textbf{0.9937} & \textbf{81.64} & \textbf{-0.32} & 165.99 & 0.1272 \\
\midrule
\multirow{7}{*}{Food-101} & \multirow{4}{*}{Latent} & Stable Diffusion & 0.8911 & 0.3383 & 0.4157 & 0.8173 & 0.1724 & 0.2882 & 156.91 & 33.83 & 161.99 & 0.9307 \\
 &  & LDCE & 0.9644 & 0.0436 & 0.0721 & 0.3174 & 0.6418 & 0.8284 & 39.44 & 7.95 & 74.58 & 0.4778 \\
 &  & DiG-IN & \underline{0.9980} & 0.0493 & 0.0793 & 0.2308 & 0.7515 & 0.7813 & 32.61 & 1.16 & 67.14 & 0.9420 \\
 &  & \mace{} (ours) & 0.9970 & \underline{0.0062} & 0.0316 & \underline{0.0406} & \underline{0.9579} & \underline{0.9888} & \textbf{29.22} & \textbf{0.06} & \textbf{61.69} & 0.1402 \\
 & \multirow{3}{*}{Pixel} & MaskDiME & 0.7703 & 0.0106 & \textbf{0.0188} & 0.0996 & 0.9288 & 0.9882 & 29.51 & 0.62 & \underline{62.21} & 0.1259 \\
 &  & RCSB & 0.9921 & 0.0130 & 0.0450 & 0.0877 & 0.9085 & 0.9713 & 30.68 & 0.57 & 65.33 & 0.2424 \\
 &  & \macesb{} (ours) & \textbf{0.9990} & \textbf{0.0033} & \underline{0.0211} & \textbf{0.0284} & \textbf{0.9725} & \textbf{0.9942} & \underline{29.45} & \underline{0.40} & 62.47 & 0.1139 \\
\midrule
\multirow{7}{*}{Oxford Pets} & \multirow{4}{*}{Latent} & Stable Diffusion & 0.9435 & 0.3202 & 0.3919 & 0.8349 & 0.1660 & 0.2779 & 132.07 & 17.32 & 142.16 & 0.8681 \\
 &  & LDCE & 0.9639 & 0.0362 & 0.0646 & 0.2617 & 0.6713 & 0.8491 & 37.25 & 6.08 & 67.02 & 0.4075 \\
 &  & DiG-IN & \underline{0.9969} & 0.0446 & 0.0805 & 0.2194 & 0.7570 & 0.7056 & 39.71 & 8.42 & 72.41 & 0.9065 \\
 &  & \mace{} (ours) & 0.9678 & \underline{0.0118} & 0.0468 & \underline{0.0638} & \underline{0.9254} & 0.9751 & 27.93 & 4.57 & 56.58 & 0.2238 \\
 & \multirow{3}{*}{Pixel} & MaskDiME & 0.7537 & 0.0121 & \underline{0.0244} & 0.0845 & 0.9184 & \underline{0.9802} & \textbf{24.99} & \textbf{1.05} & \textbf{52.57} & 0.1512 \\
 &  & RCSB & 0.9953 & \underline{0.0118} & 0.0446 & 0.0660 & 0.9198 & 0.9595 & 34.17 & 5.85 & 61.67 & 0.3500 \\
 &  & \macesb{} (ours) & \textbf{0.9992} & \textbf{0.0032} & \textbf{0.0210} & \textbf{0.0266} & \textbf{0.9732} & \textbf{0.9919} & \underline{25.08} & \underline{1.56} & \underline{52.96} & 0.1536 \\
\midrule
\multirow{7}{*}{CUB-200} & \multirow{4}{*}{Latent} & Stable Diffusion & 0.4861 & 0.2870 & 0.3495 & 0.8424 & 0.1642 & 0.2513 & 67.87 & 8.92 & 70.78 & 0.4714 \\
 &  & LDCE & 0.9722 & 0.0333 & 0.0648 & 0.2245 & 0.7398 & 0.9164 & 21.57 & 1.86 & 40.12 & 0.3470 \\
 &  & DiG-IN & 0.9957 & 0.0302 & 0.0608 & 0.1428 & 0.8456 & 0.9207 & 22.84 & 2.56 & 38.11 & 0.6789 \\
 &  & \mace{} (ours) & 0.9872 & \underline{0.0077} & 0.0387 & \underline{0.0456} & \underline{0.9547} & \underline{0.9924} & \underline{16.34} & \underline{0.55} & \textbf{31.56} & 0.2005 \\
 & \multirow{3}{*}{Pixel} & MaskDiME & 0.8779 & 0.0094 & \underline{0.0216} & 0.0740 & 0.9411 & 0.9866 & 17.27 & 0.58 & 33.95 & 0.1949 \\
 &  & RCSB & \textbf{1.0000} & 0.0126 & 0.0495 & 0.0741 & 0.9210 & 0.9671 & 24.52 & 2.68 & 42.67 & 0.3378 \\
 &  & \macesb{} (ours) & \underline{0.9995} & \textbf{0.0028} & \textbf{0.0208} & \textbf{0.0249} & \textbf{0.9792} & \textbf{0.9952} & \textbf{16.00} & \textbf{0.34} & \underline{32.03} & 0.1699 \\
\bottomrule
\end{tabular}}
\caption{Quantitative comparison of counterfactual generation methods with SigLIP ViT-B/16 as the explained model on the full benchmark (972/1010/1275/1868 instances for ImageNet/Food-101/Oxford Pets/CUB-200). We report target-class validity, proximity and realism metrics and cross-model validity (CMV) over five held-out CLIP models. All metrics except validity and CMV are computed on the common valid set (482/696/865/740 instances for ImageNet/Food-101/Oxford Pets/CUB-200). KID is reported in units of $10^{-3}$. Bold and underlined values indicate the best and second-best results, respectively, within each dataset and metric column. Gen. indicates the diffusion model type (latent or pixel space).}
\label{tab:appendix_target_siglip}
\end{table*}

\begin{table*}[t]
    \centering
    \small
    \setlength{\tabcolsep}{3pt}
\begin{tabular}{lllcccccccc}
\toprule
Explained & Variant & Attribution & Validity $\uparrow$ & $\ell_1^{\text{norm}}\downarrow$ & LPIPS $\downarrow$ & SSIM $\uparrow$ & FID $\downarrow$ & KID $\downarrow$ & sFID $\downarrow$ & CMV \\
\midrule
ViT-B/16 & \mace{} & CAV & 0.9992 & 0.0048 & 0.0277 & 0.9689 & 22.90 & 0.46 & 58.93 & 0.1017 \\
 &  & Grad-ECLIP & 1.0000 & 0.0046 & 0.0277 & 0.9691 & 23.07 & 0.44 & 59.17 & 0.1104 \\
 &  & M2IB & 0.9921 & 0.0047 & 0.0298 & 0.9695 & 22.65 & 0.37 & 58.82 & 0.1014 \\
 & \macesb{} & CAV & 0.9997 & 0.0034 & 0.0255 & 0.9739 & 23.19 & 0.51 & 59.24 & 0.1182 \\
 &  & Grad-ECLIP & 0.9998 & 0.0034 & 0.0258 & 0.9737 & 23.50 & 0.58 & 59.58 & 0.1288 \\
 &  & M2IB & 0.9957 & 0.0031 & 0.0268 & 0.9756 & 22.83 & 0.44 & 59.18 & 0.1130 \\
\midrule
ViT-L/14 & \mace{} & CAV & 0.9997 & 0.0050 & 0.0289 & 0.9669 & 22.52 & 0.36 & 58.05 & 0.1224 \\
 &  & Grad-ECLIP & 0.9996 & 0.0046 & 0.0276 & 0.9684 & 22.70 & 0.44 & 58.24 & 0.1173 \\
 & \macesb{} & CAV & 0.9993 & 0.0035 & 0.0249 & 0.9727 & 22.94 & 0.68 & 57.73 & 0.1265 \\
 &  & Grad-ECLIP & 0.9988 & 0.0033 & 0.0233 & 0.9738 & 23.15 & 0.79 & 57.90 & 0.1267 \\
\midrule
ViT-B/32 & \mace{} & CAV & 0.9997 & 0.0048 & 0.0268 & 0.9717 & 22.37 & 0.34 & 59.01 & 0.0827 \\
 &  & Grad-ECLIP & 0.9975 & 0.0043 & 0.0269 & 0.9740 & 22.54 & 0.44 & 59.00 & 0.0779 \\
 & \macesb{} & CAV & 0.9995 & 0.0034 & 0.0251 & 0.9755 & 22.84 & 0.37 & 59.63 & 0.0835 \\
 &  & Grad-ECLIP & 0.9992 & 0.0030 & 0.0247 & 0.9776 & 22.94 & 0.39 & 59.74 & 0.0805 \\
\midrule
SigLIP B/16 & \mace{} & LeGrad & 0.9703 & 0.0087 & 0.0499 & 0.9452 & 28.83 & 1.21 & 66.18 & 0.1851 \\
 &  & Grad-ECLIP & 0.8045 & 0.0082 & 0.0514 & 0.9472 & 29.14 & 1.32 & 66.48 & 0.1610 \\
 & \macesb{} & LeGrad & 0.9992 & 0.0034 & 0.0280 & 0.9734 & 23.06 & 0.13 & 58.82 & 0.1411 \\
 &  & Grad-ECLIP & 0.9618 & 0.0029 & 0.0250 & 0.9769 & 22.95 & 0.16 & 58.69 & 0.1282 \\
\bottomrule
\end{tabular}
\caption{\mace with different attribution methods, mean over the four datasets (full benchmark of each dataset). Validity and CMV are computed over all instances; proximity, FID, KID and sFID on the common valid set of the respective comparison (instances for which all attribution methods of a variant yield a valid counterfactual). KID is reported in units of $10^{-3}$; FID and sFID depend on the number of instances and are only comparable within a block.}
\label{tab:appendix_attribution}
\end{table*}

\begin{table*}[tp]
    \centering
    \small
    \setlength{\tabcolsep}{4pt}
\begin{tabular}{lccccc}
\toprule
Mask & Validity $\uparrow$ & $\ell_1^{\text{norm}}\downarrow$ & LPIPS $\downarrow$ & SSIM $\uparrow$ & Area (\%) \\
\midrule
Fixed $r=0.05$ & 0.883 & 0.0044 & 0.026 & 0.971 & 4.7 \\
Fixed $r=0.1$ & 0.970 & 0.0076 & 0.039 & 0.950 & 8.4 \\
Fixed $r=0.15$ & 0.992 & 0.0108 & 0.053 & 0.929 & 11.9 \\
Fixed $r=0.2$ & 0.995 & 0.0140 & 0.067 & 0.910 & 15.3 \\
Fixed $r=0.25$ & 0.995 & 0.0170 & 0.078 & 0.892 & 18.5 \\
Fixed $r=0.3$ & 0.998 & 0.0201 & 0.091 & 0.873 & 21.7 \\
Fixed $r=0.4$ & 0.998 & 0.0259 & 0.112 & 0.839 & 27.8 \\
Fixed $r=0.5$ & 1.000 & 0.0314 & 0.134 & 0.806 & 33.6 \\
Fixed $r=0.6$ & 1.000 & 0.0368 & 0.152 & 0.776 & 39.3 \\
Fixed $r=0.7$ & 1.000 & 0.0424 & 0.171 & 0.746 & 44.9 \\
Fixed $r=0.8$ & 1.000 & 0.0474 & 0.188 & 0.718 & 50.1 \\
Fixed $r=1$ & 1.000 & 0.0578 & 0.221 & 0.667 & 59.9 \\
\midrule
\textbf{Adaptive (ours)} & \textbf{1.000} & \textbf{0.0044} & \textbf{0.026} & \textbf{0.971} & \textbf{5.1} \\
\bottomrule
\end{tabular}
\caption{\mace{} with the counterfactual mask fixed to the top $r$ fraction of $R_{\mathrm{diff}}$ for all evaluated mask sizes, compared with the adaptive search (100 instances per dataset, mean over the four datasets; extends Table~2b of the main paper). Validity is computed over all instances, proximity metrics on the instances that all settings change to the target. Area: final mask area as a percentage of the image.}
\label{tab:appendix_fixed_mask}
\end{table*}

\begin{figure*}[p]
    \centering
    \includegraphics[width=\textwidth]{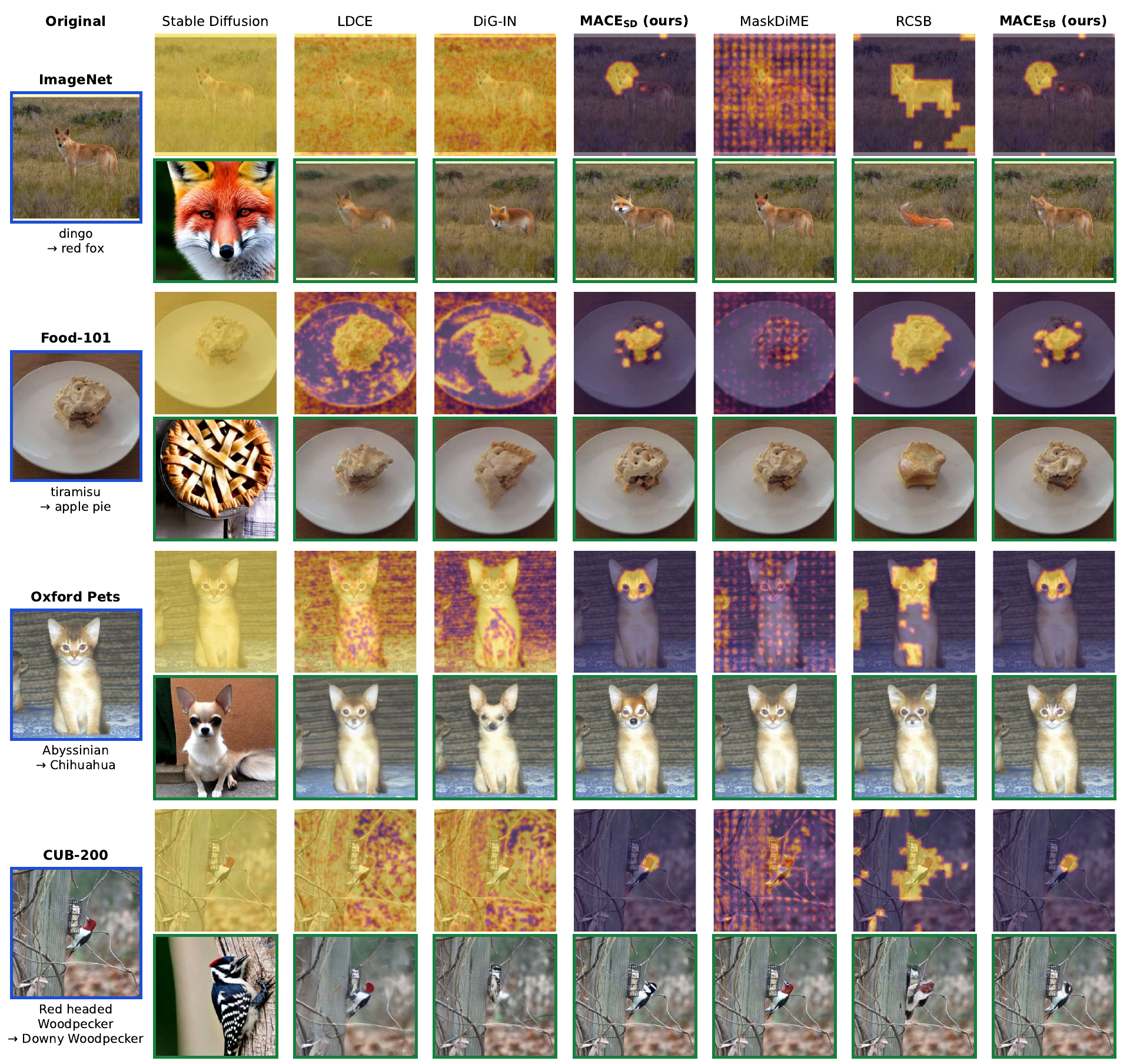}
    \caption{Additional qualitative comparison on instances with lower-ranked targets selected where \mace{} performs well, one per dataset (target: the class ranked 6th by CLIP for ImageNet, Food-101, and CUB-200, and 11th for Oxford Pets). Layout as in Figure~2 of the main paper: the original image on the left (blue frame); for each method, the top row highlights the pixels that changed (bright), and the bottom row shows the counterfactual. A green frame marks a counterfactual whose top-1 class under the explained CLIP ViT-B/16 is the target class, a red frame one that does not reach the target.}
    \label{fig:appendix_qual_a}
\end{figure*}

\begin{figure*}[t]
    \centering
    \includegraphics[width=\linewidth]{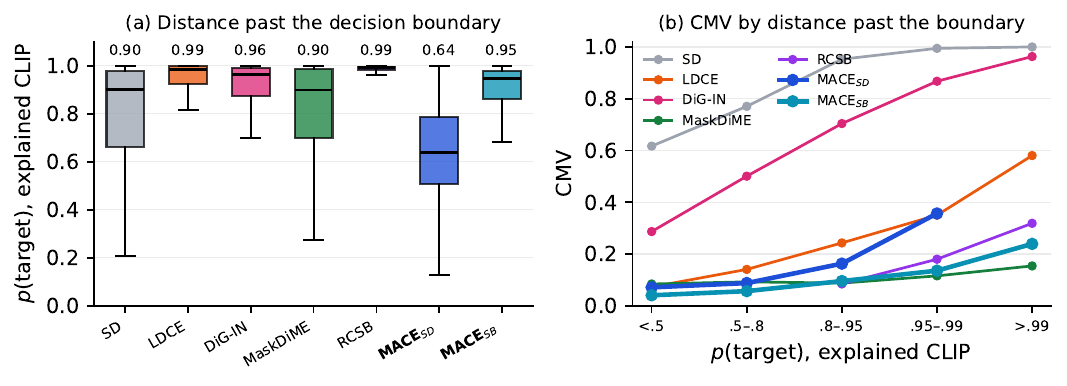}
    \caption{Distance past the decision boundary on the full benchmark (valid counterfactuals, pooled over the four datasets). (a) Target probability under the explained CLIP (ViT-B/16); numbers show the median. (b) CMV per bin of the target probability under the explained CLIP; bins with at least 30 counterfactuals.}
    \label{fig:appendix_boundary}
\end{figure*}

\begin{figure*}[tp]
    \centering
    \includegraphics[width=0.5\textwidth]{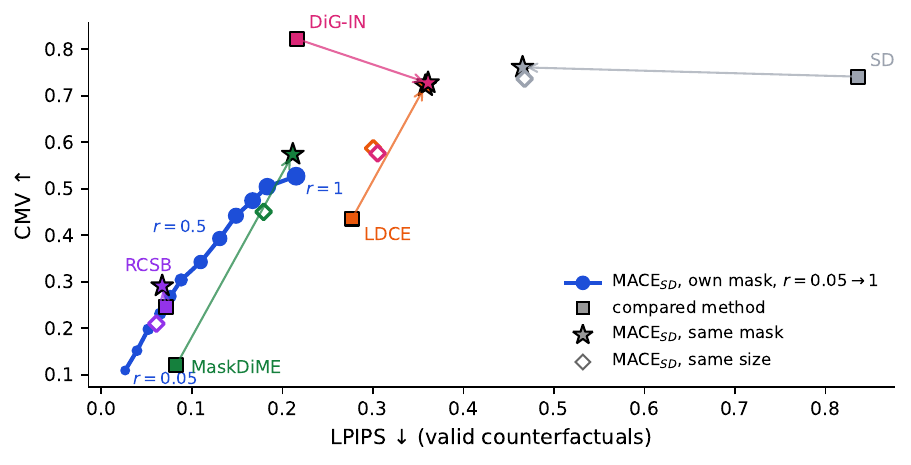}
    \caption{CMV versus LPIPS as in Figure~3b of the main paper, with \mace{} using its own counterfactual mask at the size of each compared method's changed region added (diamonds).}
    \label{fig:appendix_same_size}
\end{figure*}

\begin{table*}[t]
    \centering
    \small
    \setlength{\tabcolsep}{3pt}
\begin{tabular}{lccccccc}
\toprule
 & & \multicolumn{2}{c}{Compared method} & \multicolumn{2}{c}{\mace{}, same mask} & \multicolumn{2}{c}{\mace{}, same size} \\
\cmidrule(lr){3-4}\cmidrule(lr){5-6}\cmidrule(lr){7-8}
Region of & Area (\%) & CMV $\uparrow$ & LPIPS $\downarrow$ & CMV $\uparrow$ & LPIPS $\downarrow$ & CMV $\uparrow$ & LPIPS $\downarrow$ \\
\midrule
SD & 97.1 & 0.741 & 0.836 & 0.761 & 0.465 & 0.736 & 0.468 \\
LDCE & 72.0 & 0.436 & 0.277 & 0.722 & 0.358 & 0.587 & 0.300 \\
DiG-IN & 73.5 & 0.822 & 0.216 & 0.728 & 0.361 & 0.576 & 0.305 \\
MaskDiME & 45.2 & 0.121 & 0.083 & 0.574 & 0.212 & 0.450 & 0.179 \\
RCSB & 14.3 & 0.246 & 0.071 & 0.291 & 0.067 & 0.210 & 0.061 \\
\bottomrule
\end{tabular}
\caption{\mace{} restricted to each compared method's changed region (same mask) or using its own counterfactual mask at the size of that region (same size), without adaptive mask expansion (100 instances per dataset, mean over the four datasets). Area: mean area of the compared method's changed region. CMV over all instances; LPIPS on the instances with valid counterfactuals in all settings of a region.}
\label{tab:appendix_same_size}
\end{table*}

\begin{figure*}[p]
    \centering
    \includegraphics[width=\textwidth]{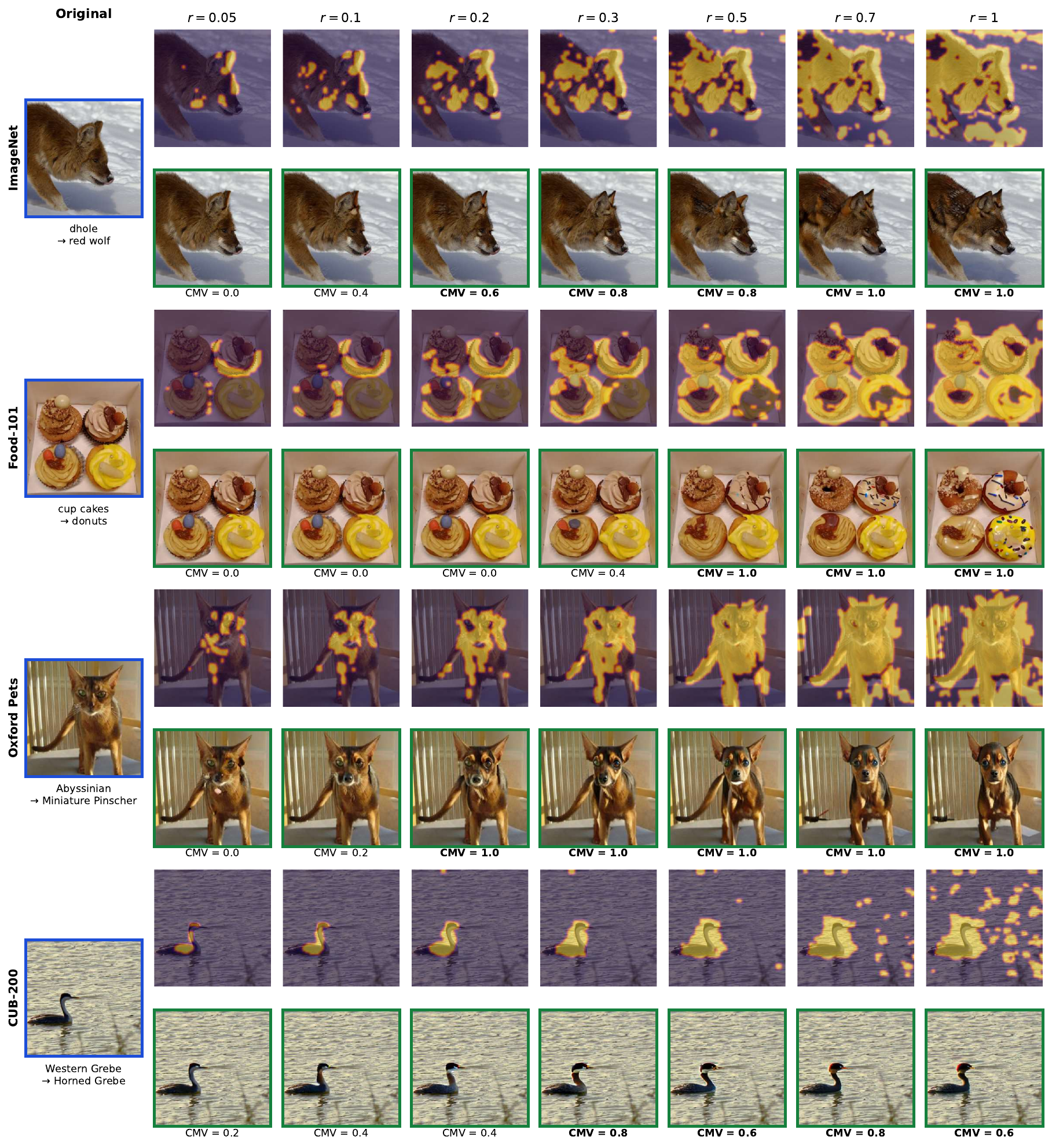}
    \caption{\mace{} counterfactuals with increasing mask size $r$ (top $r$ fraction of $R_{\mathrm{diff}}$, own generation per size) for one selected instance per dataset. For each dataset, the top row shows the editable mask over the original image and the bottom row the counterfactual. A green frame marks a counterfactual that is valid under the explained CLIP (ViT-B/16). CMV: share of the five held-out CLIPs that predict the target.}
    \label{fig:appendix_size_examples}
\end{figure*}

\begin{figure*}[p]
    \centering
    \includegraphics[width=\textwidth]{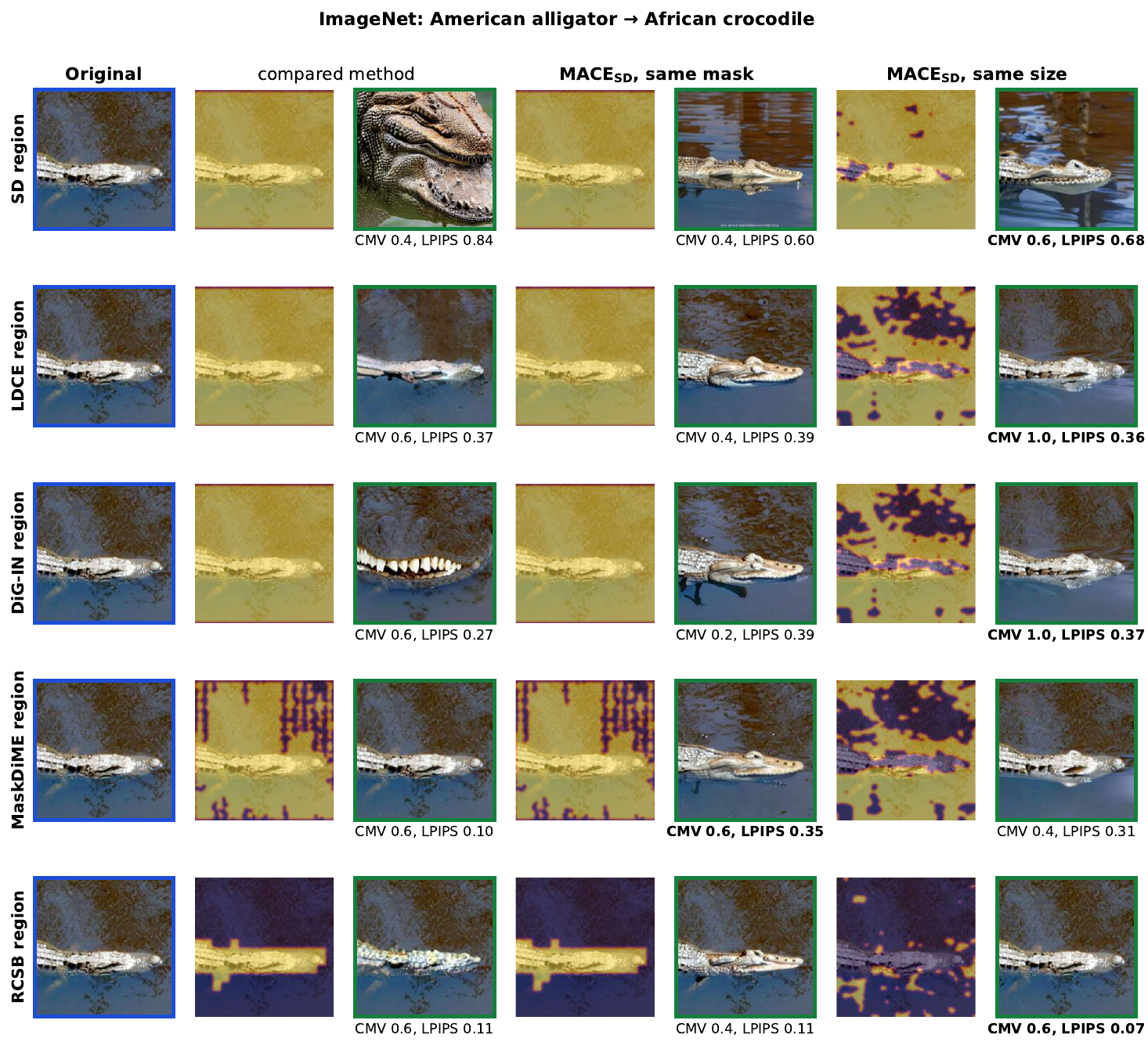}
    \caption{Same mask and same mask size on a randomly selected ImageNet instance. Each row uses the region changed by one compared method. Left to right: the original (blue frame); the compared method's changed region and its counterfactual; \mace{} restricted to exactly this region (same mask); and \mace{} with its own CAV mask at the size of this region (same size). A green frame marks a counterfactual whose top-1 class under the explained CLIP ViT-B/16 is the target, a red frame one that does not reach the target. Below each counterfactual: CMV and LPIPS.}
    \label{fig:appendix_region_imagenet}
\end{figure*}

\begin{figure*}[p]
    \centering
    \includegraphics[width=\textwidth]{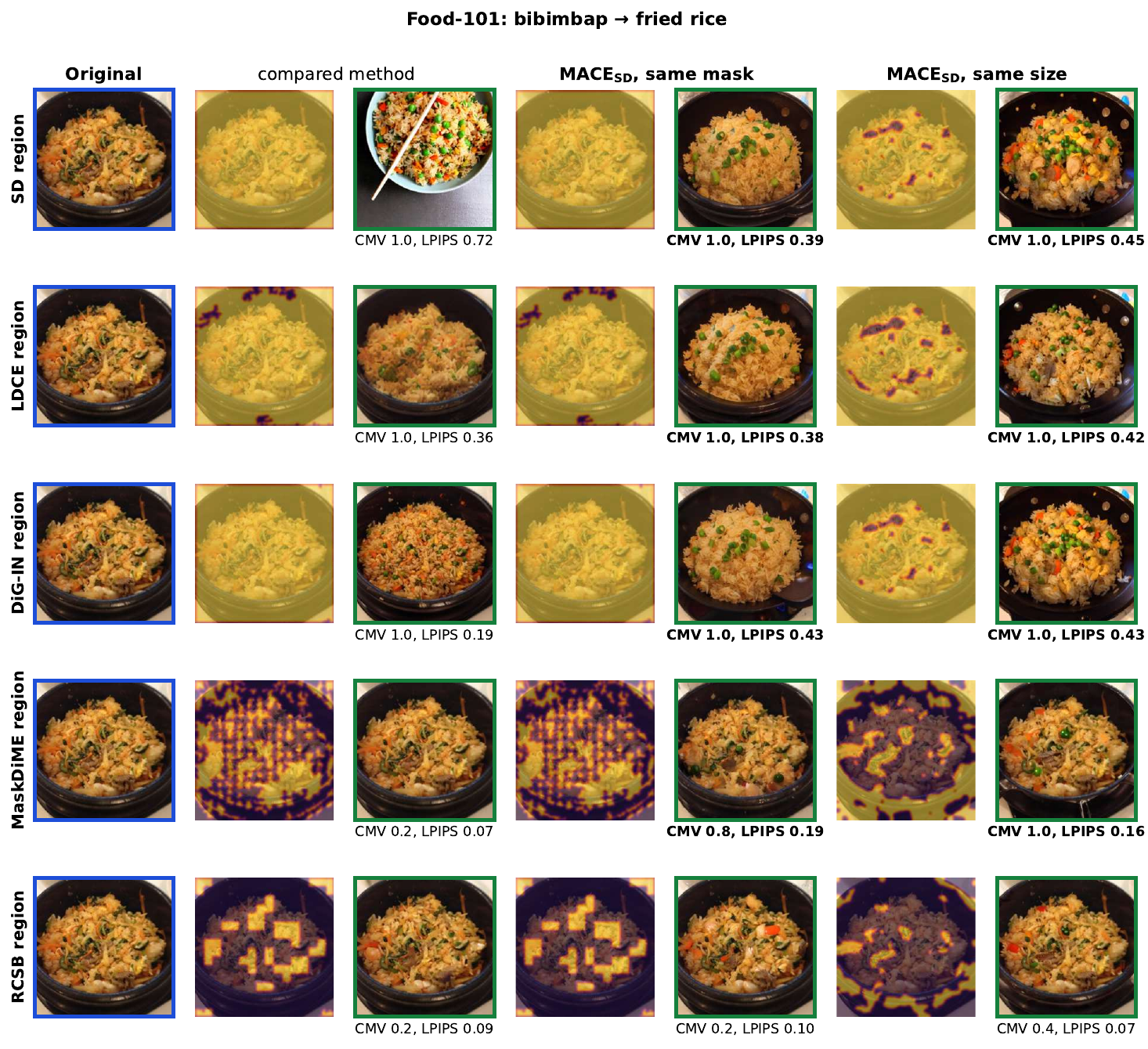}
    \caption{Same mask and same mask size on a randomly selected Food-101 instance. Each row uses the region changed by one compared method. Left to right: the original (blue frame); the compared method's changed region and its counterfactual; \mace{} restricted to exactly this region (same mask); and \mace{} with its own CAV mask at the size of this region (same size). A green frame marks a counterfactual whose top-1 class under the explained CLIP ViT-B/16 is the target, a red frame one that does not reach the target. Below each counterfactual: CMV and LPIPS.}
    \label{fig:appendix_region_food101}
\end{figure*}

\begin{figure*}[p]
    \centering
    \includegraphics[width=\textwidth]{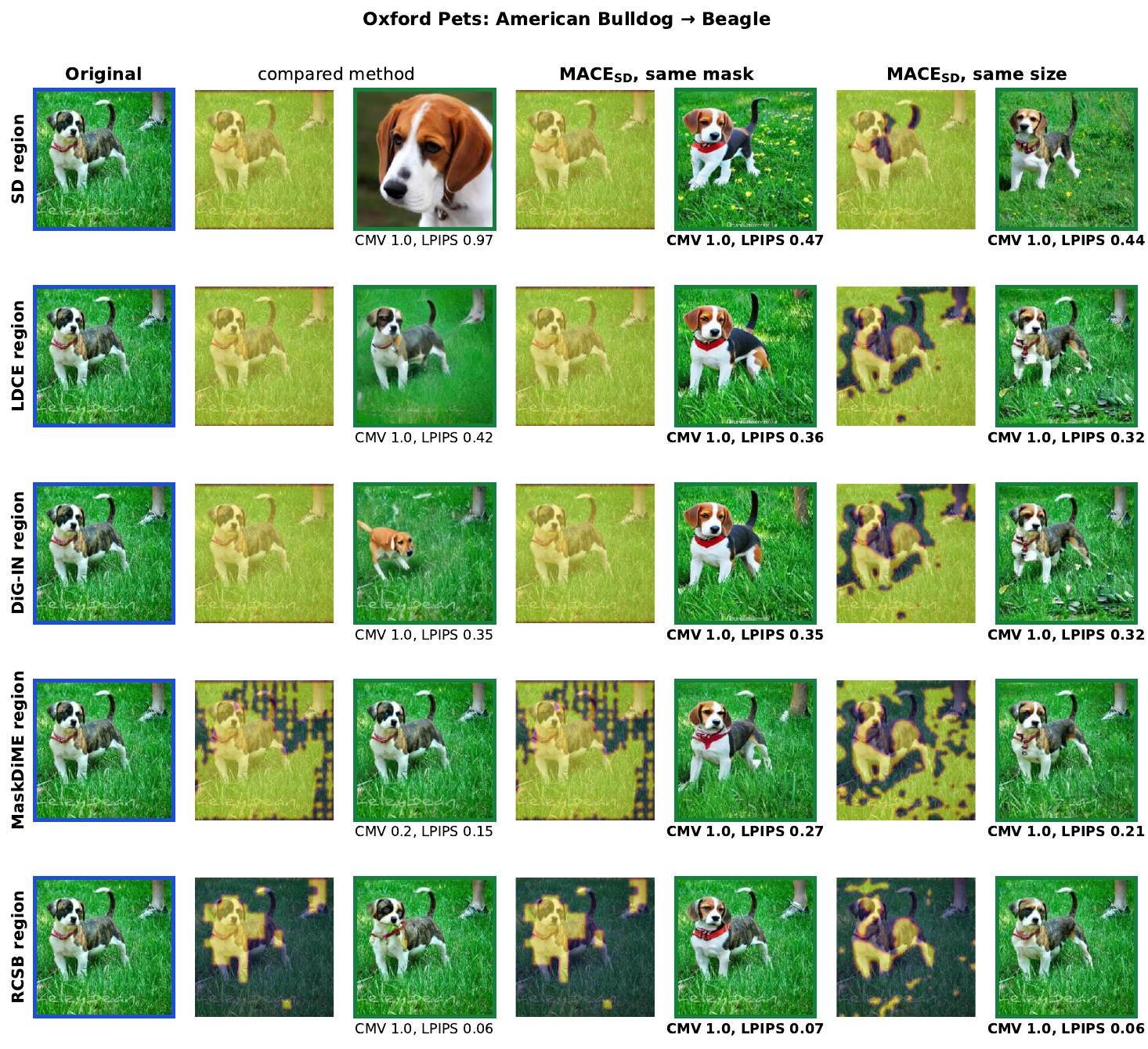}
    \caption{Same mask and same mask size on a randomly selected Oxford Pets instance. Each row uses the region changed by one compared method. Left to right: the original (blue frame); the compared method's changed region and its counterfactual; \mace{} restricted to exactly this region (same mask); and \mace{} with its own CAV mask at the size of this region (same size). A green frame marks a counterfactual whose top-1 class under the explained CLIP ViT-B/16 is the target, a red frame one that does not reach the target. Below each counterfactual: CMV and LPIPS.}
    \label{fig:appendix_region_oxford_pets}
\end{figure*}

\begin{figure*}[p]
    \centering
    \includegraphics[width=\textwidth]{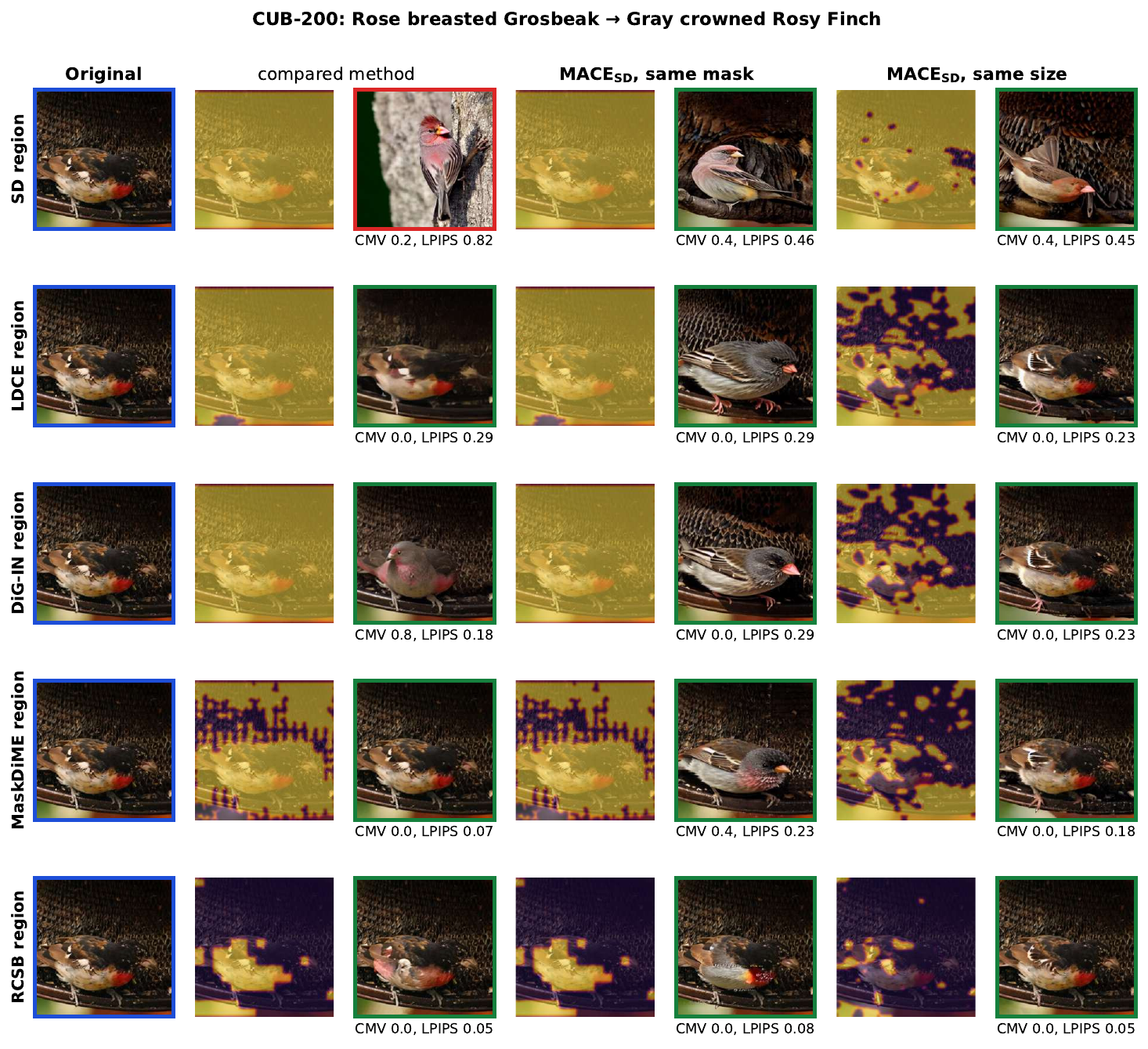}
    \caption{Same mask and same mask size on a randomly selected CUB-200 instance. Each row uses the region changed by one compared method. Left to right: the original (blue frame); the compared method's changed region and its counterfactual; \mace{} restricted to exactly this region (same mask); and \mace{} with its own CAV mask at the size of this region (same size). A green frame marks a counterfactual whose top-1 class under the explained CLIP ViT-B/16 is the target, a red frame one that does not reach the target. Below each counterfactual: CMV and LPIPS.}
    \label{fig:appendix_region_cub200}
\end{figure*}

\begin{table*}[t]
    \centering
    \small
    \setlength{\tabcolsep}{3pt}
\resizebox{\textwidth}{!}{%
\begin{tabular}{llcccccccccc}
\toprule
 & & Validity & \multicolumn{5}{c}{Proximity} & \multicolumn{3}{c}{Realism} & \\
\cmidrule(lr){3-3}\cmidrule(lr){4-8}\cmidrule(lr){9-11}
Dataset & Guidance step & Target $\uparrow$ & $\ell_1^{\text{norm}}\downarrow$ & $\ell_2^{\text{norm}}\downarrow$ & LPIPS $\downarrow$ & SSIM $\uparrow$ & \sss{} $\uparrow$ & FID $\downarrow$ & KID $\downarrow$ & sFID $\downarrow$ & CMV $\uparrow$ \\
\midrule
\multirow{2}{*}{ImageNet} & Adam (default) & 0.9989 & \textbf{0.0045} & \textbf{0.0305} & \textbf{0.0240} & \textbf{0.9707} & \textbf{0.9938} & \textbf{75.61} & \textbf{0.29} & 159.12 & 0.0959 \\
 & $\ell_2$ step & 0.9989 & 0.0051 & 0.0367 & 0.0309 & 0.9698 & 0.9908 & 76.59 & 0.95 & \textbf{158.93} & \textbf{0.1317} \\
\midrule
\multirow{2}{*}{Food-101} & Adam (default) & 0.9990 & \textbf{0.0044} & \textbf{0.0274} & \textbf{0.0308} & \textbf{0.9685} & \textbf{0.9919} & \textbf{24.45} & \textbf{1.30} & \textbf{55.41} & 0.0899 \\
 & $\ell_2$ step & \textbf{1.0000} & 0.0051 & 0.0334 & 0.0384 & 0.9668 & 0.9875 & 25.25 & 1.65 & 56.37 & \textbf{0.1370} \\
\midrule
\multirow{2}{*}{Oxford Pets} & Adam (default) & 1.0000 & \textbf{0.0044} & \textbf{0.0300} & \textbf{0.0248} & \textbf{0.9707} & \textbf{0.9901} & \textbf{25.38} & \textbf{1.11} & \textbf{48.82} & 0.0926 \\
 & $\ell_2$ step & 1.0000 & 0.0052 & 0.0370 & 0.0312 & 0.9705 & 0.9862 & 28.09 & 2.35 & 52.05 & \textbf{0.1528} \\
\midrule
\multirow{2}{*}{CUB-200} & Adam (default) & 1.0000 & \textbf{0.0045} & \textbf{0.0313} & \textbf{0.0230} & \textbf{0.9732} & \textbf{0.9959} & \textbf{15.56} & \textbf{1.00} & \textbf{31.27} & 0.1381 \\
 & $\ell_2$ step & 1.0000 & 0.0049 & 0.0356 & 0.0271 & 0.9727 & 0.9936 & 17.47 & 1.85 & 32.58 & \textbf{0.1964} \\
\midrule
\multirow{2}{*}{Mean} & Adam (default) & 0.9995 & \textbf{0.0045} & \textbf{0.0298} & \textbf{0.0257} & \textbf{0.9708} & \textbf{0.9929} & \textbf{35.25} & \textbf{0.92} & \textbf{73.66} & 0.1041 \\
 & $\ell_2$ step & \textbf{0.9997} & 0.0051 & 0.0357 & 0.0319 & 0.9700 & 0.9895 & 36.85 & 1.70 & 74.98 & \textbf{0.1545} \\
\bottomrule
\end{tabular}%
}
\caption{\mace{} with the Adam-normalized guidance step (default) and with the $\ell_2$-normalized step without momentum, on the full benchmarks. Validity and CMV are computed over all instances; proximity, \sss{}, FID, KID, and sFID on the common valid set of Table~\ref{tab:main_results}. KID is reported in units of $10^{-3}$. Bold values indicate the better variant.}
\label{tab:appendix_l2}
\end{table*}

\begin{table*}[t]
    \centering
    \small
    \setlength{\tabcolsep}{3pt}
\begin{tabular}{lccccccccc}
\toprule
 & & \multicolumn{4}{c}{Same mask} & \multicolumn{4}{c}{Same size} \\
\cmidrule(lr){3-6}\cmidrule(lr){7-10}
 & & \multicolumn{2}{c}{Adam step} & \multicolumn{2}{c}{$\ell_2$ step} & \multicolumn{2}{c}{Adam step} & \multicolumn{2}{c}{$\ell_2$ step} \\
\cmidrule(lr){3-4}\cmidrule(lr){5-6}\cmidrule(lr){7-8}\cmidrule(lr){9-10}
Region of & Area (\%) & CMV $\uparrow$ & LPIPS $\downarrow$ & CMV $\uparrow$ & LPIPS $\downarrow$ & CMV $\uparrow$ & LPIPS $\downarrow$ & CMV $\uparrow$ & LPIPS $\downarrow$ \\
\midrule
Stable Diffusion & 97.1 & 0.761 & 0.465 & 0.831 & 0.458 & 0.736 & 0.468 & 0.804 & 0.457 \\
LDCE & 72.0 & 0.722 & 0.358 & 0.786 & 0.355 & 0.587 & 0.300 & 0.632 & 0.293 \\
DiG-IN & 73.5 & 0.728 & 0.361 & 0.792 & 0.360 & 0.576 & 0.305 & 0.651 & 0.294 \\
MaskDiME & 45.2 & 0.574 & 0.212 & 0.659 & 0.211 & 0.451 & 0.179 & 0.519 & 0.177 \\
RCSB & 14.3 & 0.291 & 0.067 & 0.401 & 0.072 & 0.210 & 0.061 & 0.304 & 0.065 \\
\bottomrule
\end{tabular}
\caption{\mace{} with the Adam-normalized and the $\ell_2$-normalized guidance step, restricted to the changed region of each compared method (same mask) or using its own counterfactual mask at the size of that region (same size), without adaptive mask expansion (100 instances per dataset, mean over the four datasets). Area: mean area of the compared method's changed region. CMV over all instances; LPIPS on the instances with valid counterfactuals in all settings of a region.}
\label{tab:appendix_l2_region}
\end{table*}

\clearpage
\makeatletter
\ifdefined\supponly
    \clearpage
    \makeatletter
    \if@filesw
      \immediate\write\@auxout{\string\bibstyle{ieeenat_fullname}}%
      \immediate\write\@auxout{\string\bibdata{main}}%
    \fi
    \def\suppsetnum#1{\global\c@NAT@ctr\numexpr#1-1\relax}
    \makeatother
    {
        \small
        \renewcommand\bibsection{\section*{References for the Supplementary Material}}%
        \InputIfFileExists{supp_refs.bbl}{}{}
    }
\fi

\else
\maketitle
\begin{abstract}
Existing CLIP explanations identify regions important to the model's predictions but do not show how to change them to obtain a different prediction. Visual counterfactual methods can be adapted to CLIP but do not exploit its vision-language representation. We introduce \textbf{M}ask-guided \textbf{A}daptive \textbf{C}ounterfactual \textbf{E}xplanations (\mace), a method for generating targeted visual counterfactuals for CLIP zero-shot classification. \mace uses source-target attribution differences and adaptive mask expansion to determine \textbf{where} to edit, and CLIP-guided diffusion inpainting to determine \textbf{how} to edit.
We evaluate \mace against strong existing visual counterfactual methods adapted to CLIP zero-shot classification. Across four datasets, multiple CLIP models, and both latent- and pixel-space diffusion generators, \mace achieves near-perfect target-class validity with smaller edits than competing methods.
We further show that \mace's small edits often produce counterfactuals specific to the explained model, while larger editable regions increase cross-model validity. By controlling the editable region, \mace can thus navigate the trade-off between small, model-specific edits and more transferable counterfactuals.
The source code is available at \url{https://anonymous.4open.science/r/MACE-04BC/}.
\end{abstract}
    
\section{Introduction}
\label{sec:intro}

\begin{figure*}[t]
    \centering
    \includegraphics[width=\linewidth]{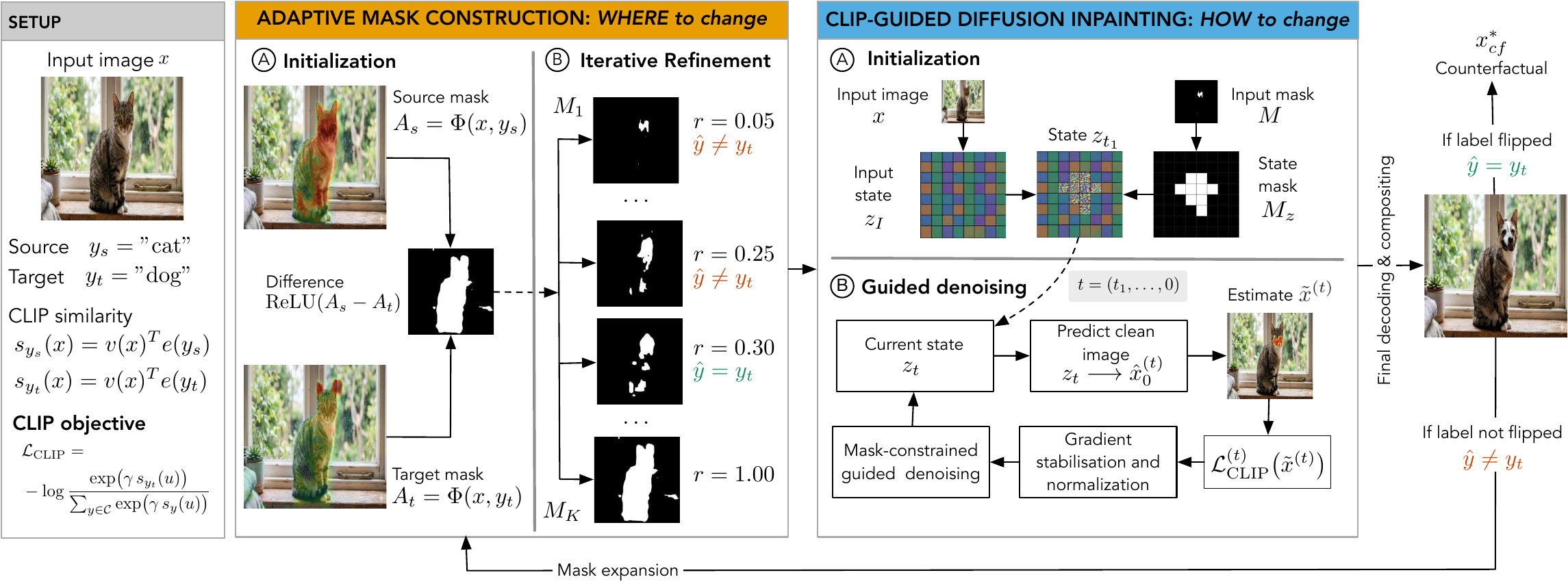}
    \caption{Overview of \mace. Given a source image and a target class, CLIP attribution identifies regions that favor the source over the target. Starting with the top attribution values ($M_1$), \mace progressively expands the resulting editable mask in 5\% steps until a candidate reaches the target class and uses CLIP-guided generative inpainting to inpaint only the selected region while preserving the remaining image content.}
    \label{fig:method-overview}
\end{figure*}

CLIP~\cite{radford2021clip} is widely used for zero-shot classification~\cite{sammani2024interpreting,martin_transductive_2024,qian_online_2024}, retrieval~\cite{Qin_2025_CVPR,xie_ra-clip_2023,sain_clip_2023}, and multimodal applications ranging from safety inspection~\cite{tsai2025construction,poppi_safe-clip_2025} to decision support~\cite{zhao_clip_2025}. However, CLIP explanations remain dominated by attention and gradient maps, which highlight relevant regions without showing how the input must change to alter the prediction~\cite{li2024clipsurgery,zhao2024gradient,luu2025visual}. Counterfactual explanations address this limitation by identifying a small, meaningful change that produces a different model outcome~\cite{wachter2017counterfactual}.

Existing visual counterfactual methods mainly target conventional image classifiers~\cite{goyal2019counterfactual,boreiko2022sparse,augustin2022diffusion}. More recent methods, such as LDCE~\cite{schrodi2024ldce}, RCSB~\cite{sobieski2024rethinking}, and DiG-IN~\cite{augustin2024digin}, use classifier gradients to guide image generation and can therefore be adapted to CLIP. However, they use CLIP only through its class logits, like a conventional classifier, and ignore the structure of its zero-shot classification, where each class is defined by a text prompt and predictions depend on image-text similarities across all candidate classes. This structure allows CLIP-specific attribution methods~\cite{chen2025class,zhao2024gradient,wang2024m2ib} to identify regions that support the source and target class prompts. CLIP-guided editing methods~\cite{yu2022cfclip,gal2021stylegannada} use text prompts to guide generation, but increasing similarity to a target prompt does not guarantee that the target becomes the top prediction among all candidate classes. This motivates a counterfactual method that builds on this structure to reliably change CLIP's zero-shot prediction while keeping edits small and localized.

We therefore propose \textbf{M}ask-guided \textbf{A}daptive \textbf{C}ounterfactual \textbf{E}xplanations (\mace), a visual counterfactual method designed specifically for targeted explanations of CLIP zero-shot predictions (overview in \cref{fig:method-overview}). \mace uses the difference between source and target CLIP attributions to identify \textit{where} to edit, then guides a generative diffusion inpainting model toward the target class using CLIP guidance to determine \textit{how} to edit. 
Starting with a small editable region mask, \mace expands this mask only when the target label is not reached. It accepts the first successful candidate and preserves the original pixels outside the mask. 
\mace is training-free, operating directly on the pretrained CLIP and diffusion models (without any fine-tuning) to find small, localized changes that are sufficient to change CLIP's prediction.

 Across ImageNet, Food-101, Oxford Pets, and CUB-200, \mace achieves almost perfect validity while making smaller edits than all other compared methods adapted to CLIP. These results also hold across different CLIP variants. 
 Our cross-model transferability analysis shows that \mace identifies model-specific counterfactuals: small image edits that are sufficient to change the \emph{explained} model's prediction, but not the prediction of \emph{other} models. Allowing larger editable regions enables transfer across models, recovering the prediction change also in other models.

In summary, our main contributions are:
\begin{itemize}[leftmargin=*]
    \item We introduce \mace, a training-free framework for targeted visual counterfactual explanations of CLIP zero-shot predictions, combining CLIP-guided diffusion inpainting with localized image editing.

    \item We propose an adaptive mask construction strategy that uses source-target attribution differences to identify relevant regions and expands the editable mask only when needed. Ablation experiments demonstrate that both components help achieve high validity with small edits.
    
    \item We demonstrate, across four datasets, different CLIP variants, and latent- and pixel-space diffusion generators that \mace achieves near-perfect target-class validity with smaller edits than strong existing counterfactual methods adapted to CLIP.
    
    \item We analyze the model specificity of \mace's counterfactuals and show how varying the editable region allows \mace to navigate the trade-off between small, model-specific edits and higher cross-model validity.
\end{itemize}

\section{Related Work}
\label{sec:related-work}

Our work connects visual counterfactual explanations with vision-language model interpretability. We review visual counterfactual generation, CLIP explainability, CLIP-guided image editing, and counterfactual learning for vision-language models, and then position \mace.

\subsection{Visual Counterfactual Explanations}
\label{subsec:RW:VCE}

Visual counterfactual explanations show how an image can be changed to obtain a target prediction with minimal changes to the original image. Early work~\citep{goyal2019counterfactual,boreiko2022sparse,thiagarajan2021designing} uses region replacement~\citep{goyal2019counterfactual}, sparse optimization~\citep{boreiko2022sparse}, and deep model inversion~\citep{thiagarajan2021designing} to generate visual counterfactuals with small, realistic changes. More recent diffusion-based methods, including DiME~\citep{jeanneret2022dime}, DVCE~\citep{augustin2022diffusion}, and LDCE~\citep{schrodi2024ldce}, guide generation using classifier objectives. Other methods improve spatial control through object-level editing~\citep{zemni2023octet}, region constraints~\citep{sobieski2024rethinking}, and adaptive masking~\citep{guo2026maskdime}, or support black-box classifiers~\citep{jeanneret2024time}. DiG-IN~\citep{augustin2024digin} provides a framework for explaining classifiers through diffusion guidance. These methods can be adapted to CLIP, but rely only on its class scores without explicitly exploiting its shared image-text embedding space. In contrast, CLIP-specific attribution methods can leverage this structure to identify image regions associated with individual class prompts.

\subsection{CLIP Explainability}

CLIP~\cite{radford2021clip} has mainly been interpreted using attribution methods~\citep{zhao2024gradient,bousselham2025legrad}, visual localization~\citep{li2024clipsurgery,chen2025class}, concept-based explanations~\citep{bhalla2024splice}, and multimodal information bottlenecks~\citep{wang2024m2ib}. These methods identify regions or concepts important to CLIP's predictions but do not show how changing them would produce a different (zero-shot) prediction, i.e., they do not provide counterfactual explanations. 

\subsection{CLIP-Guided Image Editing}
\label{subsec:RW:CGIE}
Besides counterfactual explanations, CLIP has also been used to guide image editing through text descriptions. StyleCLIP~\cite{patashnik2021styleclip} and StyleGAN-NADA~\cite{gal2021stylegannada} optimize latent directions or generator parameters to align generated images with target text descriptions. DiffusionCLIP~\citep{kim2022diffusionclip} extends CLIP-guided editing to diffusion models, enabling text-driven edits of real images while preserving their visual identity. CF-CLIP~\citep{yu2022cfclip} introduces contrastive objectives and text-guided latent optimization to produce more accurate edits. However, these methods either require additional training or fine-tuning, or rely on domain-specific pretrained generators, making them difficult to apply directly across diverse image classification datasets. Furthermore, these methods aim to match target descriptions, without guarantees that the target becomes the top-ranked class. They do not produce counterfactuals that reliably flip CLIP’s zero-shot prediction with small, localized changes.

\subsection{Counterfactual Learning for VLMs}

Counterfactuals are also used to improve the training of VLMs. Counterfactual Prompt Learning~\citep{he2022cpl} uses changes in feature representations to improve prompt generalization, while COMO~\citep{lai2024como} and CF-VLM~\citep{zhang2025cfvlm} use counterfactual examples to improve compositional understanding and sensitivity to semantic changes. Unlike our work, these methods focus on improving model learning rather than generating image-level explanations for individual predictions of a pretrained model.

\subsection{Positioning of Our Work}
Unlike existing counterfactual methods, \mace uses CLIP-specific attribution~\citep{chen2025class,zhao2024gradient,wang2024m2ib} to identify regions that support the source more than the target. Unlike CLIP explainability methods, it shows how these regions can change, and unlike CLIP-guided editing, it needs neither additional training nor a domain-specific generator, and requires the target to become CLIP’s top prediction. \mace thus answers a key question: \emph{What small, localized change is sufficient to make CLIP predict the target class instead of the source class?} Finding such changes can help uncover unexpected sensitivities of the explained model and provide insights into its decisions.
\section{Problem Formulation}
\label{sec:problem}

Our goal is to generate targeted visual counterfactual explanations for CLIP-based zero-shot image classification. Let \(\mathcal{X}\subseteq[0,1]^{H\times W\times 3}\) denote the image space, where each image \(x\in\mathcal{X}\) has three color channels and spatial resolution \(H\times W\). Let \(\mathcal{C}\) be the set of semantic classes and \(\mathcal{T}\) the space of textual prompts. CLIP~\citep{radford2021clip} consists of an image encoder \(F_{\mathrm{I}}\) and a text encoder \(F_{\mathrm{T}}\).

For each class \(c \in \mathcal{C}\), we define a text prompt \(t_c \in \mathcal{T}\) and compute the normalized image and text embeddings
\[
\bm{v}(x)=\frac{F_{\mathrm{I}}(x)}{\lVert F_{\mathrm{I}}(x)\rVert_2}, \qquad \bm{e}(t_c)=\frac{F_{\mathrm{T}}(t_c)}{\lVert F_{\mathrm{T}}(t_c)\rVert_2}.
\] The zero-shot CLIP classifier scores each class using cosine similarity
$
s_c(x)=\bigl\langle \bm{v}(x),\bm{e}(t_c)\bigr\rangle
$
and predicts
\begin{equation}
\hat{y}(x)=\arg\max_{c\in\mathcal{C}}s_c(x).
\label{eq:clip-prediction}
\end{equation}

Given an input image \(x\in\mathcal{X}\), let \(y_s=\hat{y}(x)\) denote its source prediction and let \(y_t\in\mathcal{C}\setminus\{y_s\}\) be a desired target class. Following standard counterfactual desiderata~\citep{wachter2017counterfactual,goyal2019counterfactual}, our objective is to generate a counterfactual image \(x_{\mathrm{cf}}\) that is classified as \(y_t\), remains close to the original image \(x\), and is visually realistic.

Let \(d:\mathcal{X}\times\mathcal{X}\to\mathbb{R}^{+}\)  be an image distance measure and \(\mathcal{X}_{\mathrm{real}}\subseteq\mathcal{X}\) the set of visually realistic images. We define the feasible set as

\begin{equation}
\mathcal{F}_{\mathrm{cf}}(y_t)=\left\{x'\in\mathcal{X}_{\mathrm{real}}\;\middle|\;\hat{y}(x')=y_t\right\}.
\end{equation}

The ideal counterfactual is then

\begin{equation}
x_{\mathrm{cf}}^*=\operatorname*{arg\,min}_{x'\in\mathcal{F}_{\mathrm{cf}}(y_t)}d(x,x').
\end{equation}

\section{Method}
\label{sec:method}

Mask-guided Adaptive Counterfactual Explanations (\mace), illustrated in Figure~\ref{fig:method-overview}, first determines \emph{where} to edit the image (\cref{sec:adaptive-mask}) and then \emph{how} to edit it, such that the CLIP classifier predicts the target class (\cref{sec:counterfactual-inpainting}).

\subsection{Adaptive Counterfactual Mask Construction}
\label{sec:adaptive-mask}

Let $\Phi(x,s_y)=A_y\in\mathbb{R}^{H\times W}$ denote a label-conditioned attribution operator, where $A_y(p)$ measures the contribution of pixel $p$ to the CLIP score $s_y(x)$. To identify where to edit for a transition from source class $y_s$ to target class $y_t$, we propose
a \textit{contrastive relevance map} based on the attribution maps $A_s=\Phi(x,s_{y_s})$ and $A_t=\Phi(x,s_{y_t})$:
\begin{equation}
R_{\mathrm{diff}}=\operatorname{ReLU}(A_s-A_t).
\label{eq:contrastive-relevance}
\end{equation}
Unlike a standard attribution map that highlights features important to the current prediction, $R_{\mathrm{diff}}$ identifies regions that support the source more than the target. It assigns low scores to regions that support both classes similarly or already favor the target, helping preserve these features, and high scores to regions where changes may be needed to reach the target.

\paragraph{Iterative Refinement.}
Not all regions identified by $R_{\mathrm{diff}}$ need to change to achieve the target prediction. To find a counterfactual with minimal changes, we start with the regions with the highest scores and expand the editable area only when generation fails to reach the target class.

At attempt $k\in\{1,\dots,K\}$, we construct a binary \textit{counterfactual mask} $M_k\in\{0,1\}^{H\times W}$ from $R_{\mathrm{diff}}$ using a quantile threshold:
\begin{equation*}
M_k(p)=\mathbb{I}\left[
R_{\mathrm{diff}}(p)\geq
Q_{1-r_k}(R_{\mathrm{diff}})
\right],
\end{equation*}
where $M_k(p)=1$ marks an editable pixel, $r_k\in(0,1]$ controls the fraction of pixels selected, and $Q_{1-r_k}$ is the corresponding quantile across all pixels.

As $r_k$ increases, the mask keeps previously selected pixels and adds regions with lower difference scores. We stop expanding the mask when a candidate reaches the target prediction or after $K$ attempts.

For each mask $M_k$, we generate a candidate $x_{\mathrm{cf}}^{(k)}$ and accept the first one that achieves the target prediction, $\hat{y}(x_{\mathrm{cf}}^{(k)})=y_t$. If all attempts fail, we select the candidate with the largest margin between the target score and the highest score among the other classes:

\begin{equation*}
k^{\star}=\arg\max_k\left[s_{y_t}\!\left(x_{\mathrm{cf}}^{(k)}\right)-\max_{c\in\mathcal{C}\setminus\{y_t\}}s_c\!\left(x_{\mathrm{cf}}^{(k)}\right)\right].
\end{equation*}

\subsection{CLIP-Guided Generative Inpainting}
\label{sec:counterfactual-inpainting}

The mask $M$ determines \textit{where} to edit. We use a pretrained inpainting model to determine \textit{how} to edit through iterative denoising. Let $z_t\in\mathbb R^{h\times w\times C_z}$ be the denoising state at timestep $t$, which decreases from a starting step $t_1\le T$ to $0$, $z_I$ the input state, i.e., the input $x$ in the representation of the generative model, and $z_I^{(t)}$ its reference state at the same timestep. We adapt $M$ to the dimensions of $z_t$ to obtain $M_z$. After each denoising step, locations with $M_z=0$ are reset to $z_I^{(t)}$, so only locations with $M_z=1$ can change.

\paragraph{CLIP guidance.}
At each denoising step, we want to steer the generation toward the target class using CLIP guidance. However, the current state $z_t$ is still noisy, so CLIP cannot judge it directly. As in forward universal guidance~\citep{bansal2024universal}, we instead let the denoiser predict the clean state $\hat z_0(z_t)$, i.e., its estimate of the final state $z_0$ given $z_t$, and evaluate CLIP on the corresponding image $\hat x_0^{(t)}=D(\hat z_0(z_t))$, where $D$ maps a state to an image with the same resolution and value range as $x$. Unlike universal guidance, but similar to Blended Diffusion~\citep{avrahami2022blended}, we combine the predicted clean image inside the mask with the original image outside the mask:
\begin{equation}
\widetilde x^{(t)}=M\odot \hat x_0^{(t)}+(1-M)\odot x.
\label{eq:guidance-image}
\end{equation}
By restoring the regions outside $M$, we let CLIP evaluate an estimate of the counterfactual image, with changes restricted to the masked region.

To encourage CLIP to assign the highest score to the target class $y_t$ over all other classes, we guide image generation using a cross-entropy loss computed on the estimated counterfactual $\widetilde x^{(t)}$:
\begin{equation}
\mathcal L_{\mathrm{CLIP}}^{(t)}=-\log
\frac{\exp\!\big(\gamma\, s_{y_t}(\widetilde x^{(t)})\big)}
{\sum_{y\in\mathcal C}\exp\!\big(\gamma\, s_y(\widetilde x^{(t)})\big)},
\label{eq:clip-ce}
\end{equation}
where $s_y$ is the CLIP score of \cref{eq:clip-prediction} and $\gamma$ is CLIP's logit scale. Its negative gradient $g_t=-\nabla_{z_t}\mathcal L_{\mathrm{CLIP}}^{(t)}$, backpropagated through CLIP, $D$, and the denoiser, provides the direction for updating the noisy state to increase CLIP's confidence in the target class.

\paragraph{Guided update and preservation.}
Because CLIP gradients can fluctuate across denoising steps, we follow \citet{sobieski2024rethinking} and use Adam~\citep{kingma2015adam} to stabilize the CLIP gradient $g_t$, obtaining $\hat g_t$, and normalize it by the norm of the first stabilized gradient $\hat g_{t_1}$:
\begin{equation}
\kappa=\max\{\|\hat g_{t_1}\|_2,\epsilon_\kappa\},
\qquad \epsilon_\kappa>0.
\label{eq:stabilization}
\end{equation}
We incorporate the stabilized gradient into the denoising update:
\begin{equation}
\bar z_{t-1}=\mathcal S_t(z_t,q_t)+\lambda w_t\frac{\hat g_t}{\kappa},
\label{eq:update}
\end{equation}
where $\mathcal S_t$ is the deterministic scheduler, $q_t$ is the denoiser output, $\lambda$ controls the guidance strength, and $w_t$ is a model-specific weight. To preserve the unmasked regions, we restore them using the corresponding noisy input $z_I^{(t-1)}$ at the next timestep $t-1$:
\begin{equation}
z_{t-1}=M_z\odot\bar z_{t-1}+(1-M_z)\odot z_I^{(t-1)}.
\label{eq:restore}
\end{equation}

After the final denoising step, we construct the counterfactual image as
\begin{equation*}
x_{\mathrm{cf}}=M\odot D(z_0)+(1-M)\odot x.
\end{equation*}
This keeps every pixel outside $M$ unchanged.


\begin{table*}[t]
    \centering
    \tiny
    \setlength{\tabcolsep}{3pt}
\resizebox{\textwidth}{!}{
\begin{tabular}{lllccccccccc}
\toprule
 & & & \multirow{2}{*}{Validity $\uparrow$}
 & \multicolumn{5}{c}{Proximity}
 & \multicolumn{3}{c}{Realism} \\
\cmidrule(lr){5-9}\cmidrule(lr){10-12}
Dataset & Gen. & Method/Variant &
&
$\ell_1^{\text{norm}}\downarrow$ &
$\ell_2^{\text{norm}}\downarrow$ &
LPIPS $\downarrow$ &
SSIM $\uparrow$ &
\sss{} $\uparrow$ &
FID $\downarrow$ &
KID $\downarrow$ &
sFID $\downarrow$
\\
\midrule
\multirow{6}{*}{ImageNet}
 & Pixel & MaskDiME & 0.7859 & 0.0133 & \textbf{0.0287} & 0.1009 & 0.9068 & \underline{0.9797} & \underline{76.07} & \textbf{0.01} & 161.50 \\
 & Pixel & RCSB & \underline{0.9957} & \underline{0.0118} & 0.0460 & \underline{0.0751} & \underline{0.9191} & 0.9698 & 78.67 & 1.56 & 162.06 \\ & Latent & Stable Diffusion & 0.7195 & 0.3337 & 0.4050 & 0.8378 & 0.1327 & 0.2165 & 102.03 & 4.87 & 160.37 \\
 & Latent & LDCE & 0.9711 & 0.0514 & 0.0919 & 0.3263 & 0.6064 & 0.8266 & 84.81 & 6.30 & 166.13 \\
 & Latent & DiG-IN & 0.9604 & 0.0547 & 0.0974 & 0.2765 & 0.7051 & 0.6571 & 78.29 & 4.91 & \textbf{154.22} \\
 & Latent & \textbf{\mace{} (ours)} & \textbf{0.9989} & \textbf{0.0045} & \underline{0.0306} & \textbf{0.0240} & \textbf{0.9707} & \textbf{0.9938} & \textbf{75.61} & \underline{0.29} & \underline{159.28} \\
\midrule
\multirow{6}{*}{Food-101}
 & Pixel & MaskDiME & 0.8993 & \underline{0.0094} & \textbf{0.0178} & 0.0871 & \underline{0.9349} & \underline{0.9840} & \underline{24.54} & \textbf{1.15} & \underline{56.41} \\
 & Pixel & RCSB & \textbf{1.0000} & 0.0120 & 0.0425 & \underline{0.0798} & 0.9143 & 0.9746 & 25.44 & 1.77 & 57.70 \\
 & Latent & Stable Diffusion & 0.9701 & 0.3418 & 0.4199 & 0.8188 & 0.1751 & 0.2875 & 149.58 & 31.54 & 155.60 \\
 & Latent & LDCE & 0.9930 & 0.0436 & 0.0724 & 0.3134 & 0.6420 & 0.8238 & 33.96 & 8.60 & 66.82 \\
 & Latent & DiG-IN & 0.9930 & 0.0517 & 0.0834 & 0.2449 & 0.7427 & 0.7300 & 28.37 & \underline{1.24} & 62.78 \\
 & Latent & \textbf{\mace{} (ours)} & \underline{0.9990} & \textbf{0.0044} & \underline{0.0274} & \textbf{0.0308} & \textbf{0.9685} & \textbf{0.9919} & \textbf{24.46} & 1.30 & \textbf{55.85} \\
\midrule
\multirow{6}{*}{Oxford Pets}
 & Pixel & MaskDiME & 0.8157 & 0.0115 & \textbf{0.0243} & 0.0775 & 0.9203 & \underline{0.9825} & \textbf{24.80} & \underline{1.52} & \textbf{49.91} \\
 & Pixel & RCSB & \underline{0.9975} & \underline{0.0112} & 0.0427 & \underline{0.0610} & \underline{0.9218} & 0.9631 & 32.11 & 3.31 & 58.11 \\
 & Latent & Stable Diffusion & 0.8970 & 0.3186 & 0.3906 & 0.8369 & 0.1656 & 0.2856 & 131.65 & 20.14 & 135.99 \\
 & Latent & LDCE & 0.9791 & 0.0377 & 0.0671 & 0.2697 & 0.6606 & 0.8530 & 38.90 & 6.77 & 66.27 \\
 & Latent & DiG-IN & 0.9841 & 0.0463 & 0.0836 & 0.2240 & 0.7492 & 0.6941 & 38.29 & 5.70 & 66.70 \\
 & Latent & \textbf{\mace{} (ours)} & \textbf{1.0000} & \textbf{0.0044} & \underline{0.0300} & \textbf{0.0248} & \textbf{0.9707} & \textbf{0.9901} & \underline{25.38} & \textbf{1.11} & \underline{50.60} \\
\midrule
\multirow{6}{*}{CUB-200}
 & Pixel & MaskDiME & 0.9254 & \underline{0.0090} & \textbf{0.0216} & \underline{0.0668} & \underline{0.9435} & \underline{0.9890} & \underline{16.78} & \textbf{1.00} & \underline{33.69} \\
 & Pixel & RCSB & \textbf{1.0000} & 0.0122 & 0.0485 & 0.0719 & 0.9223 & 0.9680 & 25.42 & 2.42 & 43.87 \\
 & Latent & Stable Diffusion & 0.5034 & 0.2856 & 0.3481 & 0.8363 & 0.1763 & 0.3012 & 63.86 & 11.68 & 66.74 \\
 & Latent & LDCE & 0.9900 & 0.0328 & 0.0645 & 0.2186 & 0.7475 & 0.9185 & 21.89 & 2.14 & 39.94 \\
 & Latent & DiG-IN & 0.9843 & 0.0310 & 0.0630 & 0.1463 & 0.8453 & 0.9240 & 22.42 & 3.20 & 38.43 \\
 & Latent & \textbf{\mace{} (ours)} & \textbf{1.0000} & \textbf{0.0045} & \underline{0.0313} & \textbf{0.0230} & \textbf{0.9732} & \textbf{0.9959} & \textbf{15.56} & \textbf{1.00} & \textbf{31.79} \\
\bottomrule
\end{tabular}
}
\caption{Quantitative comparison of counterfactual generation methods. We report target-class validity, proximity and
realism metrics. All metrics except validity are
computed on the common valid set. KID is reported
in units of $10^{-3}$. Bold and underlined values
indicate the best and second-best results, respectively,
within each dataset and metric column. Gen. indicates the diffusion model type (latent or pixel space). Results for our pixel-space variant \macesb{} are provided in Appendix~\ref{sec:appendix_macesb}.}
\label{tab:main_results}
\end{table*}

\begin{figure*}[!t]
    \centering
    \includegraphics[width=\textwidth,height=0.57\textheight,keepaspectratio]{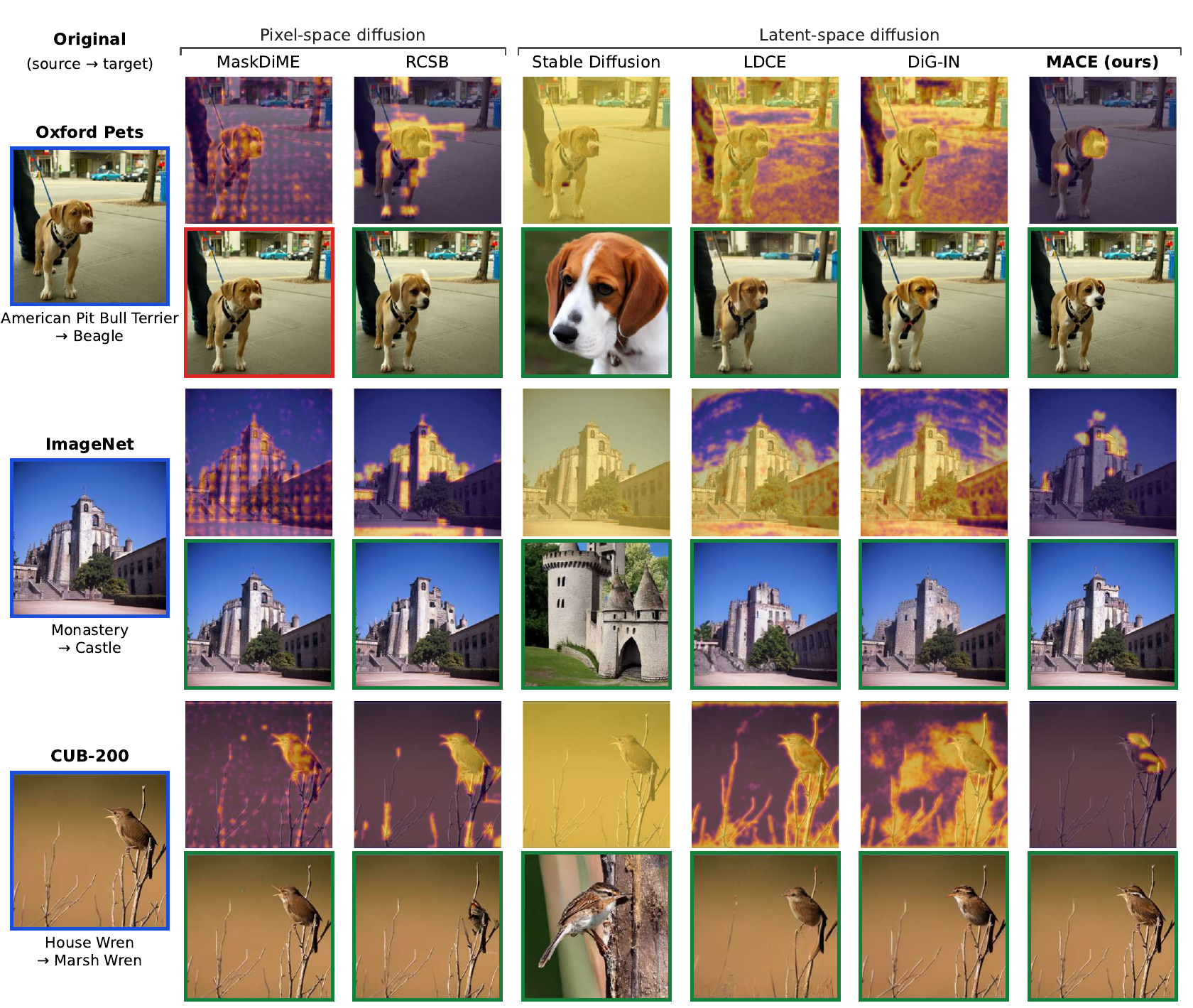}
    \caption{Qualitative comparison of all counterfactual methods on one example each from Oxford Pets (target rank = 6), ImageNet, and CUB-200 (both target rank = 2). The original image is shown on the left (blue frame) with source $\rightarrow$ target labels, where source is CLIP’s original prediction. For each method, the top row highlights the pixels that changed (bright), and the bottom row shows the counterfactual. For \mace, the changed regions also correspond to the editable masks. A green frame marks a counterfactual whose top-1 class under the explained zero-shot CLIP ViT-B/16 is the target class, a red frame one that does not reach the target.}
    \label{fig:qualitative_all_datasets}
\end{figure*}

\subsection{Generative Models}
\label{sec:generative-models}
The procedure above only requires a generative model that can inpaint a masked region, in either latent or pixel space. Our main implementation, \mace{}, uses Stable Diffusion inpainting~\cite{rombach2022latent}\footnote{\label{fn:sd-inpainting}\url{https://huggingface.co/stable-diffusion-v1-5/stable-diffusion-inpainting}}, which operates in latent space. Here, $z_t$ is a VAE latent, $D$ is the VAE decoder, $\mathcal S_t$ is the DDIM step, and $z_I^{(t)}$ is the forward-noised input latent. The denoiser is also conditioned on a target text prompt.

To demonstrate the flexibility of \mace{}, we additionally implement a pixel-space variant, \macesb{}, using the image-to-image Schr\"odinger bridge I$^2$SB~\cite{liu2023i2sb}. \macesb{} operates directly in pixel space without text conditioning. We provide its implementation details of both variants in Appendix~\ref{sec:method:backbones}. Unless otherwise specified, we use \mace{} for all
experiments in the main paper.
\section{Experimental Setup}
\label{sec:experimental_setup}
We evaluate \mace on four image classification datasets with multiple CLIP variants as explained models, comparing it with existing visual counterfactual methods adapted to the zero-shot setting on validity, proximity, and realism. This section outlines the datasets, model configurations, compared methods, and evaluation metrics.

\subsection{Datasets}
\label{sec:datasets}

We evaluate on ImageNet-1k~\cite{deng2009imagenet}, Food-101~\cite{bossard2014food101}, Oxford-IIIT Pet~\cite{parkhi2012catsdogs}, and CUB-200-2011~\cite{wah2011cub} using the ImageNet validation split and the official test splits of the other three datasets. We include only images correctly classified by zero-shot CLIP, i.e., the true class has the highest similarity among the prompts of all class names of the dataset. While several prior ImageNet evaluations of visual counterfactuals focus on three predefined class pairs~\citep{jeanneret2023ace,sobieski2024rethinking}, we select images from a broad range of source classes (1,258 in total) and limit the number of images per class to keep the computational cost manageable (Appendix~\ref{sec:appendix_dataset_sampling}). In the main experiments, we use CLIP's second-ranked class as the target, with additional results for targets ranked 6th and 11th (Appendix~\ref{sec:appendix_target_rank}). In total, the main evaluation contains 4,726 counterfactual tasks across the four datasets.

\subsection{Explained Models and Attribution}
\label{sec:configuration}

We perform our main experiments on OpenAI CLIP ViT-B/16~\cite{radford2021clip}. 
We additionally generate counterfactuals for OpenAI CLIP ViT-B/32 and ViT-L/14~\cite{radford2021clip}, and SigLIP ViT-B/16~\cite{zhai2023siglip}, evaluating all counterfactual generation methods on a subset of each dataset (Appendix~\ref{sec:appendix_target_models}). 

We compute attribution maps using CAV~\cite{chen2025class}. The adaptive mask expansion starts with the top 5\% of $R_{\mathrm{diff}}$ and expands the mask by five percentage points after each unsuccessful attempt for up to 20 attempts. We test with Grad-ECLIP~\cite{zhao2024gradient} and M2IB~\cite{wang2024m2ib} as alternative attribution methods (Appendix~\ref{sec:appendix_attribution}). The remaining hyperparameters of \mace are given in Appendix~\ref{sec:appendix_implementation}.

\subsection{Compared Methods}
\label{sec:baselines}
Most existing visual counterfactual methods are designed for supervised classifiers rather than CLIP's zero-shot setting~\cite{goyal2019counterfactual,zemni2023octet,jeanneret2024time,sobieski2024rethinking}. We therefore adapt four of them to the zero-shot setting: LDCE~\cite{schrodi2024ldce}, DiG-IN~\cite{augustin2024digin}, MaskDiME~\cite{guo2026maskdime}, and RCSB~\cite{sobieski2024rethinking}. In each method, we replace the supervised classifier with the zero-shot CLIP classifier over all class names of the dataset (details in Appendix~\ref{sec:appendix_baselines}). We also include a Stable Diffusion-only method that shares the generative model of \mace{} but uses neither attribution masks nor CLIP guidance.

\subsection{Evaluation Metrics}
\label{sec:evaluation_metrics}

We report \emph{validity} ($\uparrow$) as the fraction of counterfactuals classified as the target (target top-1 success rate)~\cite{jeanneret2023ace,schrodi2024ldce}. To assess \emph{proximity}, we measure normalized $\ell_1$ ($\downarrow$) and $\ell_2$ ($\downarrow$) distances~\cite{schrodi2024ldce}, LPIPS ($\downarrow$)~\cite{zhang2018lpips}, SSIM ($\uparrow$)~\cite{wang2004ssim}, and SimSiam similarity \sss{} ($\uparrow$)~\cite{jeanneret2023ace}. To evaluate \emph{realism}, we use Fr\'echet Inception Distance (FID, $\downarrow$)~\cite{heusel2017fid}, Kernel Inception Distance (KID, $\downarrow$)~\cite{binkowski2018kid}, and split FID (sFID, $\downarrow$)~\cite{jeanneret2023ace}, using the same reference sets across methods. Validity is computed over all instances. 
To ensure that all methods are compared on the same set of images, all other metrics are computed on the common valid set, i.e., the instances for which the counterfactuals of all methods are classified as the target.

\section{Results}
\label{sec:results}

We evaluate \mace both quantitatively and qualitatively. We examine validity, proximity, and realism across four datasets (Table~\ref{tab:main_results}), and qualitatively compare the counterfactuals generated by different methods (Figure~\ref{fig:qualitative_all_datasets}).

\subsection{Quantitative Comparison}
\label{sec:quantitative_results}

\paragraph{\mace achieves high validity with small edits across diverse datasets and models.}
Table~\ref{tab:main_results} shows that \mace achieves over 99.8\% validity on all four datasets while making the smallest edits among all compared methods on most metrics. It achieves the lowest $\ell_1$ and LPIPS and the highest SSIM and \sss{} on every dataset, with an LPIPS about 2.5 to 3 times lower than that of the next-best method. It also achieves the lowest FID on three datasets and the lowest sFID on two, so the small edits do not come at the cost of realism. These results hold across datasets with different levels of class detail, from general objects in ImageNet to fine-grained pet and bird categories in Oxford Pets and CUB-200. We observe similar results with other explained models (\cref{sec:configuration}), where \mace achieves 94 to 100\% validity with the smallest edits in most settings (Appendix~\ref{sec:appendix_target_models}). The findings also largely hold with alternative attribution methods, although attribution quality affects validity, particularly for SigLIP (Appendix~\ref{sec:appendix_attribution}).

\paragraph{\mace generalizes across generative models.}
The pixel-space variant \macesb{} (\cref{sec:generative-models}) confirms the results with a different generator. It performs similarly to \mace, with slightly better proximity but slightly worse realism (Appendix~\ref{sec:appendix_macesb}).

\paragraph{\mace remains effective for lower-ranked targets.}
The main evaluation uses CLIP's second-ranked class as the target. We further evaluate classes ranked 6th and 11th to test whether \mace remains effective for target classes that CLIP considers less likely. Results in Appendix~\ref{sec:appendix_target_rank} (Tables~\ref{tab:appendix_rank5} and~\ref{tab:appendix_rank10}) indicate that methods like LDCE or MaskDiME show a clear drop in validity, whereas \mace maintains near-perfect validity while consistently requiring smaller changes than the other methods. This demonstrates that \mace can reliably steer CLIP toward target classes with small modifications, even when these classes are not among CLIP's top-ranked predictions.

\begin{table}[t]
    \centering
    \footnotesize
    \setlength{\tabcolsep}{4pt}
\resizebox{\columnwidth}{!}{
\begin{tabular}{lccccc}
\toprule
Mask & Validity $\uparrow$ & $\ell_1^{\text{norm}}\downarrow$ & LPIPS $\downarrow$ & SSIM $\uparrow$ & Area (\%) \\
\midrule
\multicolumn{6}{c}{\textit{(a) Relevance map, adaptive expansion}} \\
\textbf{Difference $R_{\mathrm{diff}}$ (ours)} & \textbf{1.000} & \textbf{0.0046} & \textbf{0.028} & \textbf{0.969} & \textbf{5.1} \\
Source $A_s$ & 1.000 & 0.0080 & 0.039 & 0.949 & 8.2 \\
\midrule
\multicolumn{6}{c}{\textit{(b) Mask size, ranking by $R_{\mathrm{diff}}$}} \\
Fixed $r=0.05$ & 0.883 & 0.0044 & 0.026 & 0.971 & 4.7 \\
Fixed $r=0.1$ & 0.970 & 0.0076 & 0.039 & 0.950 & 8.4 \\
Fixed $r=0.2$ & 0.995 & 0.0140 & 0.067 & 0.910 & 15.3 \\
Fixed $r=0.5$ & 1.000 & 0.0314 & 0.134 & 0.806 & 33.6 \\
Fixed $r=1$ & 1.000 & 0.0578 & 0.221 & 0.667 & 59.9 \\
\textbf{Adaptive (ours)} & \textbf{1.000} & \textbf{0.0044} & \textbf{0.026} & \textbf{0.971} & \textbf{5.1} \\
\bottomrule
\end{tabular}}
\caption{Ablation of the counterfactual mask. (a) Comparison of relevance maps used to rank pixels during adaptive mask construction. (b) Fixed mask fraction $r$ versus adaptive expansion, both using $R_{\mathrm{diff}}$. Validity is computed over all instances, while proximity measures are evaluated on the common valid set of each block, ensuring comparability within but not across blocks. Area denotes the final mask size as a percentage of the image.}
\label{tab:mask_ablation}
\end{table}

\subsection{Qualitative Results}

Figure~\ref{fig:qualitative_all_datasets} shows examples of generated counterfactuals. Across all three inputs, \mace produces highly localized edits, consistent with the quantitative results above. In contrast, Stable Diffusion alters the entire image, while LDCE and DiG-IN introduce subtle but widespread pixel changes. 
For the dog, \mace primarily modifies the head, whereas MaskDiME introduces scattered, mostly imperceptible changes across the image but fails to produce a valid counterfactual. RCSB modifies a larger region around the dog. DiG-IN visibly recolors the face and chest while also making subtle changes across the rest of the dog and much of the street, which are more apparent in the change map than in the counterfactual image. For the building, \mace changes details near the top while preserving its overall appearance, whereas RCSB also modifies large parts of the facade. For the bird, \mace restricts its edits to the head and breast, while LDCE, DiG-IN, and RCSB also modify the surrounding branches. \textbf{These examples illustrate how \mace generates valid counterfactuals with small, localized edits.}
Further qualitative results for all four datasets are provided in Appendix~\ref{sec:appendix_qualitative}.

\section{Analysis}
\label{sec:ablation}

We further investigate how mask construction influences the counterfactuals generated by \mace and to what extent they are specific to the model being explained. We also examine the validity of \mace's counterfactuals on other CLIP models and how validity varies with the size of the mask.

\subsection{Mask Construction and Adaptation}
To evaluate the mask design, we conduct two ablation experiments on 100 instances per dataset, averaging results across the four datasets. First, we replace the contrastive relevance map $R_{\mathrm{diff}}$ with the source attribution $A_s$ while retaining adaptive expansion. This preserves full validity but increases LPIPS from 0.028 to 0.039 (\cref{tab:mask_ablation}a). 
Second, we replace adaptive expansion with fixed masks of varying sizes. Achieving full validity requires $r=0.5$, increasing LPIPS to 0.13 (\cref{tab:mask_ablation}b). \textbf{These results show that the contrastive relevance map helps select more relevant regions, while adaptive expansion avoids unnecessarily large masks.}

\begin{figure}[t]
    \centering
    \begin{subfigure}[b]{\linewidth}
        \centering
        \includegraphics[width=\linewidth]{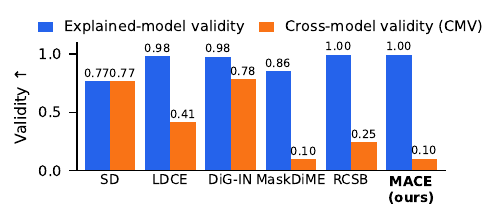}
        \caption{Explained-model validity vs.\ CMV}
        \label{fig:cmv_bar}
    \end{subfigure}\\[4pt]
    \begin{subfigure}[b]{\linewidth}
        \centering
        \includegraphics[width=\linewidth]{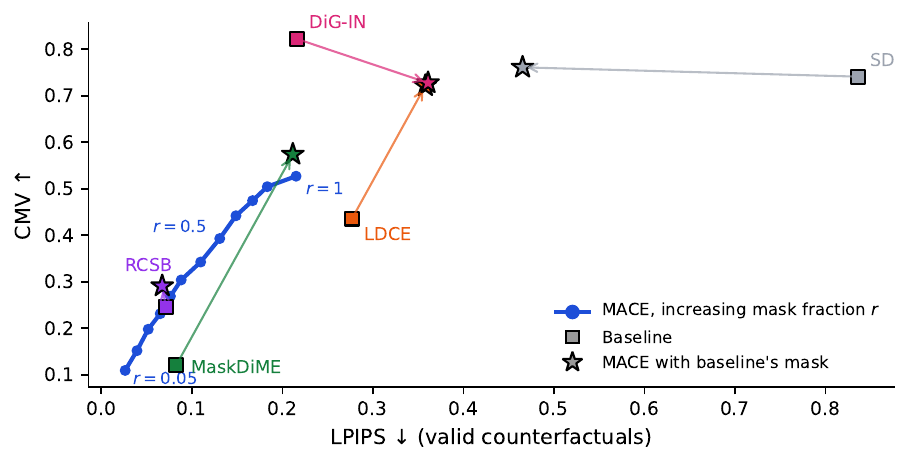}
        \caption{CMV vs.\ edit size}
        \label{fig:cmv_vs_edit}
    \end{subfigure}
    \caption{Cross-model validity (CMV). (a) Explained-model validity and CMV on the full benchmark (all instances, mean over the four datasets). 
    (b) CMV versus LPIPS on the same 100 instances per dataset. Blue curve: \mace with varying fixed mask fractions $r$. Squares: compared methods. Stars: \mace restricted to the regions modified by each compared method. Arrows connect corresponding square-star pairs.
    }
\label{fig:cmv}
\end{figure}

\subsection{Model Specificity and Cross-Model Validity}
\label{sec:cmv}
\mace finds small changes sufficient to change the explained model's prediction to the target label. We next investigate to what extent the generated counterfactuals are specific to the explained CLIP model or generic across a set of models. Following prior work~\citep{rodriguez2021beyond,dombrowski2024diffeomorphic}, we use other CLIP models as proxies for independent observers and measure cross-model validity (CMV). Prior studies suggest that cross-model agreement can align with semantic image regions~\citep{li2021crossmodel} and that plausible counterfactuals can be more robust across models~\citep{pawelczyk2020counterfactual}. While CMV does not directly measure semantic meaningfulness, it may indicate that the edits affect shared, decision-relevant features, potentially reflecting semantic changes.

We evaluate the counterfactuals on five held-out models: OpenAI CLIP ViT-L/14~\citep{radford2021clip}, OpenCLIP~\citep{cherti2023reproducible} ViT-B/16, ViT-H/14, and ConvNeXt-Base-W~\citep{liu2022convnet} trained on LAION-2B~\citep{schuhmann2022laion5b,cherti2023reproducible}, and SigLIP ViT-B/16~\citep{zhai2023siglip}. CMV is the fraction of counterfactuals classified as the target, averaged over the held-out models:
\begin{equation}
\mathrm{CMV}=\frac{1}{|\mathcal F'|}\sum_{f'\in\mathcal F'}\frac{1}{N}\sum_{i=1}^{N}
\mathbb{I}\!\left[\hat y_{f'}\big(x_{\mathrm{cf}}^{(i)}\big)=y_t^{(i)}\right],
\end{equation}
where $\mathcal F'$ is the set of held-out models, $N$ is the number of instances, and $\hat y_{f'}$ is the zero-shot top-1 prediction of model $f'$ using bare class-name prompts.

\paragraph{\mace's counterfactuals are specific to the explained model.}
Figure~\ref{fig:cmv_bar} shows that despite achieving near-perfect validity on the explained model, \mace's counterfactuals rarely transfer to other models. This is consistent with \mace’s objective and the counterfactual desiderata in \cref{sec:problem}, which seek small changes sufficient to achieve the explained model's target prediction. Such localized changes can help highlight decision-relevant features~\citep{goyal2019counterfactual} and identify model-specific sensitivities or unexpected behavior~\citep{augustin2024digin}. Furthermore, \mace achieves a median target probability of 0.64, compared with 0.99 for LDCE and RCSB (Appendix~\ref{sec:appendix_boundary}). This suggests that \mace changes the prediction without increasing target confidence as much as the other methods, helping identify what is sufficient to change the explained model's decision. 

\paragraph{\mace's cross-model validity increases with larger editable regions.}
While \mace's adaptive mask expansion favors small, model-specific changes, we investigate how larger editable regions affect cross-model validity. We first vary the fixed mask size, then evaluate \mace using the regions modified by each compared method as editable masks. We disable adaptive mask expansion in both experiments.

Figure~\ref{fig:cmv_vs_edit} shows that CMV increases with mask size (blue curve; examples in Appendix~\ref{sec:appendix_size_examples}). Given the same editable regions, \mace matches or exceeds the compared methods' CMV in most cases (stars versus squares). Qualitative examples (Appendix~\ref{sec:appendix_region_examples}) further suggest that larger editable regions can produce more recognizable target-class features. This raises the hypothesis that higher CMV may be associated with more human-recognizable semantic changes, although further evaluation is needed. Together, these results highlight a trade-off between small, model-specific edits and more transferable, potentially more human-recognizable counterfactuals.

\section{Conclusion}
\label{sec:conclusion}

We introduced \mace, a method for generating targeted visual counterfactual explanations for CLIP zero-shot classification. Across four datasets, \mace achieves near-perfect validity and, on most proximity metrics, smaller edits than existing visual counterfactual methods adapted to CLIP. These results demonstrate the value of exploiting CLIP-specific attribution and its zero-shot classification structure rather than treating CLIP solely through its class scores. Our analysis further shows that varying the editable region allows \mace to navigate the trade-off between small, model-specific edits and more transferable counterfactuals. Future work could explore multi-model guidance and human studies to investigate whether more transferable changes are also more recognizable to humans.

\section*{LLM Usage Statement}
We used OpenAI's ChatGPT and Anthropic's Claude for grammar correction and language editing. We also used Claude to assist with reproducing related work using the source code released by the original authors, debugging implementations, and generating code functions. All AI-generated suggestions and code were reviewed, verified, and revised by the authors.

{
    \small
    \bibliographystyle{ieeenat_fullname}
    \bibliography{main}
}
\clearpage
\appendix

\fi

\end{document}